\pdfoutput=1

\documentclass[11pt]{article}

\usepackage[final]{acl}

\usepackage{times}
\usepackage{latexsym}

\usepackage[T1]{fontenc}

\usepackage[utf8]{inputenc}

\usepackage{microtype}

\usepackage{inconsolata}

\usepackage{graphicx}
\usepackage[textsize=small]{todonotes}

\usepackage{amsmath}
\usepackage{amssymb}
\usepackage{xspace} 
\usepackage{bbm} 
\usepackage{booktabs}
\usepackage{longtable}
\usepackage{subcaption}
\usepackage{caption}
\usepackage[table]{xcolor}

\newcommand{\qt}[1]{``#1''}
\newcommand{\calC}{\mathcal{C}}
\newcommand{\yy}{\ensuremath{\mathbf{y}}\xspace}
\newcommand{\context}{\yy_{<t}}
\newcommand{\incontext}{(y \mid \yy_{<t})}
\newcommand{\samp}[1]{{p_{#1}}\xspace}
\newcommand{\padapt}[1]{p^{(#1)}}
\newcommand{\llama}{Llama-3.2-3B\xspace}
\newcommand{\bino}{Binoculars\xspace{}}
\newcommand{\Omegat}[1]{\Omega^{(#1)}_{t}}
\newcommand{\Zt}[1]{Z^{(#1)}_{t}}

\newcommand{\ATD}{ATD\xspace{}} 

\usepackage[most]{tcolorbox}
\tcbset{enhanced, breakable, boxrule=0.3pt, arc=2pt, colframe=red!25, colback=white}
\newtcolorbox{exampletext}[1]{title={#1}, left=6pt, right=6pt, top=6pt, bottom=6pt}

\definecolor{hl}{HTML}{FFF3BF}  
\definecolor{good}{HTML}{D3F9D8}
\definecolor{bad}{HTML}{FFDCE0} 

\newcommand{\hr}{\rowcolor{hl}}
\newcommand{\ok}{\rowcolor{good}}
\newcolumntype{G}{>{\columncolor{hl}}c}

\DeclareMathOperator*{\argmax}{argmax}
\DeclareMathOperator*{\argmin}{argmin}

\title{How Sampling Affects the Detectability of Machine-written texts:\\  A Comprehensive Study}

\author{
  \textbf{Matthieu Dubois\textsuperscript{1}} \and
  \textbf{François Yvon\textsuperscript{1}} \and
  \textbf{Pablo Piantanida\textsuperscript{2,3,4}}
\\
  \textsuperscript{1}Sorbonne Université, CNRS, ISIR, Paris France \\
  \textsuperscript{2} CNRS, International Laboratory on Learning Systems, Montréal, Canada \\
  \textsuperscript{3}Mila - Québec AI Institute, Montréal, Canada \\
  \textsuperscript{4}CentraleSupélec, Université Paris-Saclay, Gif-sur-Yvette, France
\\
  \small{
    \textbf{Correspondence:} \href{mailto:duboism@isir.upmc.fr}{duboism\;\textit{at}\;isir.upmc.fr}
  }
}

\begin{document}
\maketitle


\begin{abstract}
As texts generated by Large Language Models (LLMs) are ever more common and often indistinguishable from human-written content, research on automatic text detection has attracted growing attention. Many recent detectors report near-perfect accuracy, often boasting AUROC scores above 99\%. However, these claims typically assume fixed generation settings, leaving open the question of how robust such systems are to changes in decoding strategies. In this work, we systematically examine how sampling-based decoding impacts detectability, with a focus on how subtle variations in a model’s (sub)word-level distribution affect detection performance. We find that even minor adjustments to decoding parameters - such as temperature, top-p, or nucleus sampling - can severely impair detector accuracy, with AUROC dropping from near-perfect levels to 1\% in some settings. Our findings expose critical blind spots in current detection methods and emphasize the need for more comprehensive evaluation protocols. To facilitate future research, we release a large-scale dataset encompassing 37 decoding configurations, along with our code and evaluation framework \url{https://github.com/BaggerOfWords/Sampling-and-Detection}.


\end{abstract}

\section{Introduction \label{sec:intro}}
Texts generated with Large Language Models (LLMs) have become almost indiscernible from human-written samples \citep{gehrmann-etal-2019-gltr, clark-etal-2021-thats, dugan-etal-2023-roft}, raising concerns regarding their use for deception or harassment, among others misuses \citep{zellers-etal-2019-DefendingNeuralFake, crothers-etal-2023-machine, feng-etal-2024-bot}. To answer such threats, the task of Artificial Text Detection (ATD) has become increasingly popular, spurring competitions such as the Voight-Kampff Task @PAN24 \citep{bevendorff-etal-2024-authorship} or the Machine-Generated Text Detection Task @COLING25\citep{dugan-etal-2025-genai}.

\begin{figure}[t]
    \centering
    \includegraphics[width=0.95\linewidth]{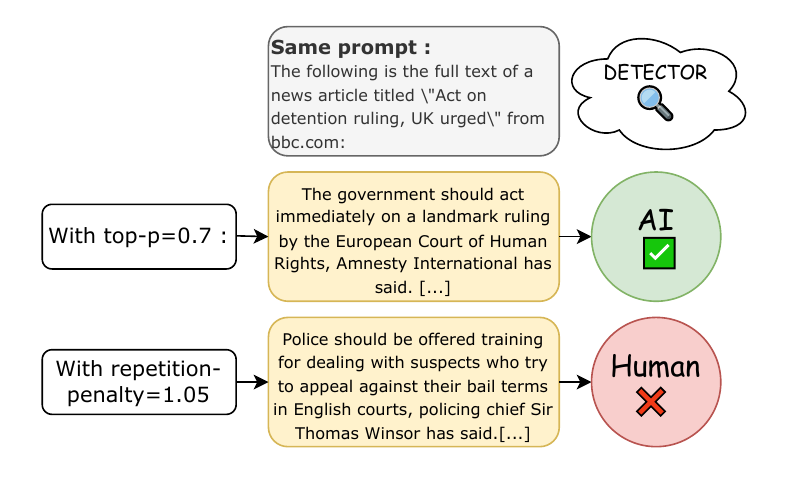}
    \label{fig:intro-image}
    \caption{Changes in the decoding strategy can lead to errors in detection.}\vspace{-4mm}
\end{figure}

Most detection systems report performance on machine-written texts generated by their authors themselves, leading to unclear comparisons and results that may not translate well to other use-cases \citep{gritsai-etal-2025-aidetectorsgoodenough}. To address this issue, large benchmarks have been developed, covering a variety of generator models \citep{li-etal-2024-mage}, languages \citep{macko-etal-2023-multitude, wang-etal-2024-m4}, domains \citep{guo-etal-2023-HowCloseChatGPT} and even attacks \citep{dugan-etal-2024-raid, wang-etal-2024-stumbling}. Among those, only \citet{dugan-etal-2024-raid} include four different sampling configurations in their RAID dataset, demonstrating that greedy outputs are easier to spot than ancestrally sampled ones, and that adding repetition penalty worsens performance in both cases.

\textbf{Our Contribution:} 
\begin{itemize}
\item We release a large-scale and diverse benchmark dataset featuring texts generated using six decoding strategies across 37 decoding configurations, enabling robust and fine-grained analysis of text detector performance.
\item Through extensive evaluation, we show that state-of-the-art detection systems are highly sensitive to generation parameters—revealing drastic performance drops (e.g., AUROC from 0.99 to 0.01)—thus exposing critical blind spots in current benchmarks.
\item We provide an in-depth analysis of these failures, uncovering the underlying mechanisms behind detection success and failure, and offering new insights into the interaction between generation dynamics and detectability.
\end{itemize}


\section{Related Work \label{sec:relatedworks}}
\textbf{Sampling Adapters}. The effects of the decoding strategy on generated text is often discussed as a trade-off between quality and diversity \citep{zhang-etal-2021-trading,meister-etal-2022-high,garces-arias-etal-2025-decoding}. \citet{meister-etal-2023-efficacy} focus on how each sampling technique modifies the token-level distribution of language models and the correlated changes in the quality of generation, depicting these alterations as a trade-off between precision and recall instead.

\textbf{Detecting artificial texts}.
Automatic detection of machine-generated text has gained attention, framed as distinguishing between human-written (“natural”) and model-generated (“artificial”) content. This generic problem can be framed in several ways: labelling texts as \qt{AI} or \qt{Human}, detecting one known artificial agent (e.g., \ ChatGPT \citep{mitrovic-etal-2023-ChatGPTHumanDetect,liu2024detectability}), or identifying one model in a predefined list \citep{liOriginTracingDetecting2023}. Other works consider finer grained labels, discriminating \qt{machine-written} from \qt{machine-polished} texts \cite{abassy-etal-2024-llm, CheckGPT}. Lastly, some datasets focus on one particular domain (e.g., scientific \citep{liyanage-etal-2022-benchmark}, academic \citep{liu2024detectability} or ``user-generated" \citep{fagni-etal-2021-TweepFakeDetectingDeepfake,kumarageStylometricDetectionAIGenerated2023}) rather than on open-domain detection.

\textbf{Detection as Text Classification.} When supervision data is available for training, artificial text detectors often achieve near-perfect accuracy rates (99\% and above) \citep{guo-etal-2023-HowCloseChatGPT,liu2024detectability,verma-etal-2023-GhostbusterDetectingText}.
Their robustness with respect to changes in the text generation model is however questioned by \citet{antoun-etal-2024-text}. In unsupervised settings, the best current methods are perturbation-based \citep{hans-etal-2024-SpottingLLMsBinoculars,mitchell-etal-2023-DetectGPT,bao-etal-2025-glimpse,dubois-etal-2025-mosaicmultipleobserversspotting} and rely on the assumption that human texts are more \qt{surprising} than machine ones. Implementing this idea typically involves to use one or several \emph{detector} model(s) to evaluate surprisal. Alternative approaches are based on text-rewriting techniques \citep{mao-etal-2024-raidar,yang-etal-2024-dnagpt}.

\textbf{Testing detectors robustness}. \ATD{} benchmarks  \citet{li-etal-2024-mage,dugan-etal-2024-raid,wang-etal-2024-stumbling} typically include multiple types of generated texts, mimicking attacks aimed at fooling detectors. 
At the generation step, varying prompts \citep{koike-etal-2024-prompt,kumarage-etal-2023-reliable,lu-etal-2024-guided} has been proven to evade detection. After generation, \textit{post-hoc} methods such as paraphrasing \citep{krishna-etal-2023-paraphrasing}, or replacing certain characters or words are very effective as well, yet they can significantly alter the original text \citep{macko-etal-2024-authorship}. 
In this work, we do not seek to deceive but simply study the influence of the decoding strategy on the detectability of the generated content, as it has been done partly in the studies of \citet{ippolito-etal-2020-automatic}, highlighting the effects of top-p and top-k sampling on detecting artificial texts, and \citet{fishchuk-etal-2023-blackbox}, tweaking the parameters of the OpenAI API to see which would evade the now discontinued OpenAI generated text detector, with frequence and presence penalty being the most impactful. We aim to provide a complete report as to how the generation strategy modifies the outputs, and why some of them go under the radar. As we observe in \textsection{}~\ref{sec:experiments}, simply tweaking the generation parameters can completely derail state-of-the-art detectors, especially the unsupervised ones.

\section{Text Generation \label{sec:language generation}}
\subsection{Language Models \label{ssec:lms}}
Language models are probability distributions over an output space $\mathcal{Y}$ containing 
all possible strings over a finite vocabulary $\Omega$:
$ 
\mathcal{Y} \triangleq \big\{ \texttt{BOS}  \circ \mathbf{y}  \circ \texttt{EOS}\,  | \,  \mathbf{y} \in \Omega^* \big\},   
$ 
where $\texttt{BOS}$ and $\texttt{EOS} $ respectively denote the beginning-of-sequence and end-of-sequence tokens and $\Omega^* $ is the Kleene closure of $\Omega$.

Neural models for text generation are parametrized with trainable weights $\theta\in\Theta$ and follow a local-normalization scheme, meaning that $\forall$ $t > 0$, $ p_\theta (\cdot | \mathbf{y}_{<t}, )$ defines a conditional  distribution over $\bar{\Omega} = \Omega \cup \texttt{EOS}$. The probability of sequence $\mathbf{y}  = \langle y_0,  \ldots, y_T \rangle$ factors as:
\begin{equation}
    p (\mathbf{y}) = \prod_{t=1}^{T} 
   p_\theta ( y_t | \mathbf{y}_{<t}),
    \label{eq:sequence_probability-eq}
\end{equation}
with $\mathbf{y}_{<t} = \langle y_0, \ldots, y_{t-1} \rangle$, $y_0 = \texttt{BOS}$; $y_T = \texttt{EOS}$. 

\subsection{Decoding Strategies \label{ssec:decoding}}
To generate text with a language model, a \emph{decoding strategy} needs to be formalized, which describes how the next token will be chosen from $p_\theta$ in order to extend the current string. Starting from a prefix $\mathbf{y}$ of length $S$, either equal to the $\langle\texttt{BOS}\rangle$ token or the user’s prompt, and at each time step $t$ we generate from
$
  p_\theta\bigl(\cdot\mid \context\bigr)\quad\text{with}\quad
  \context=\langle \mathbf{y},y_{S+1},\dots,y_{S+t-1} \rangle.
$

\textbf{Sampling Adapters}. 
Many generation strategies can be seen as sampling from an \emph{adapted distribution} derived from the model’s output \(p_\theta\), using transformation functions known as \emph{adapters}~\citep{meister-etal-2023-efficacy}:
$$
  \alpha_{\text{param}}
  :\;\Delta^{|\bar\Omega|-1}
  \;\longrightarrow\;
  \Delta^{|\bar\Omega|-1}
$$ 
that transforms the base distribution \(p_\theta(\cdot\mid\yy_{<t})\) over the full vocabulary \(\bar\Omega\) into a new distribution over a (possibly smaller) support.  We write
\begin{align}
  \padapt{\,\text{param}\,}(\cdot \mid\yy_{<t})
  \;\triangleq\;
  \alpha_{\text{param}}\bigl(p_\theta(\cdot\mid\yy_{<t})\bigr).
\end{align}

In this study, we consider six adapters, representative of the most common generation strategies.

\textbf{Temperature Sampling} is defined as:
\[
  \padapt{\mathrm{temp}=T}(y\mid\yy_{<t})
  \propto
  p_\theta(y\mid\yy_{<t})^{1/T}.
\]for some temperature parameter $T>0$. $T=1$ corresponds to \textbf{ancestral sampling}, selecting tokens using the actual model distribution; the limiting case $T\rightarrow 0$ corresponds to \textbf{greedy decoding}.

\textbf{Repetition penalty} \citep{keskar-etal-2019-ctrl} is defined as:
\[
  \padapt{\mathrm{rep}=T}(y\mid\yy_{<t})
  \hspace{-0.1cm} = \hspace{-0.15cm}
  \begin{cases}
    \displaystyle
    \frac{p_\theta(y\mid\yy_{<t})^{1/T}}{Z_{\calC}}
    , & \hspace{-0.4cm} y\in\{\yy_{<t}\},\\
    \displaystyle
    \frac{p_\theta(y\mid\yy_{<t})}{Z_{\calC}}
    , & \hspace{-0.4cm} \text{otherwise.}
  \end{cases}
\]
Where $Z_{\calC} = \sum_{y'\in\calC}p(y'\mid\context)$ is a normaliser and $T>0$ the repetition parameter. With $T>1$, this makes tokens that have been already be generated less likely, favouring diversity in the output.

Another family of popular sampling adapters rely on truncation functions, concentrating the probability mass on a strict subset of $\Omega$, and assigning a null probability to all the others tokens. This means that for adapter $\mathcal{A}$, you select a subset \(\Omegat{\mathcal A}\subseteq\bar\Omega\) at each step \(t\), then renormalize:
\[
  \padapt{\mathcal A}(y\mid\yy_{<t})
  = 
  \begin{cases}
    \displaystyle
    \frac{p_\theta(y\mid\yy_{<t})}{\Zt{\mathcal A}}
    , & y\in \Omegat{\mathcal A},\\[1ex]
    0, & \text{otherwise},
  \end{cases}
\]
where   $Zt^{(\mathcal A)} = \sum_{y'\in\Omegat{\mathcal A}} p_\theta(y'\mid\yy_{<t})$. We consider the four following truncation adapters, each associated with a specific criterion.  

\textbf{Top-p (nucleus) sampling} \cite{holtzman-etal-2020-thecurious} relies on: 
\begin{align}
  \Omegat{\mathrm{top}\text{-}p}
  &= \argmin_{\Omega' \subseteq \bar\Omega} |\Omega'|  \\ 
  &\text{where}
    \sum_{y \in \Omega'} \samp{\theta}(y \mid \yy_{<t}) \;\ge\; p. \nonumber
\end{align}
A token can only be generated if it belongs to the smallest subset of $\Omega$ whose combined probability mass is greater than $p$.

\textbf{Top-$k$ sampling} \cite{fan-etal-2018-hierarchical} relies on:
\begin{align}
    \Omegat{\mathrm{top}\text{-}k} = &\argmax_{\Omega' \subseteq \bar{\Omega}} \sum_{y \in \Omega'} \samp{\theta}(y \mid \yy_{<t}) \\
    &\,\,\text{where}\,\,|\Omega'|=k\nonumber.
\end{align}
This strategy only generates tokens that are among the $k$ most likely ones.

\textbf{Locally typical sampling} \cite{meister-etal-2023-locally} relies on:
\begin{align*}
    \Omegat{\mathrm{typical}} &= \argmin_{\Omega' \subseteq \bar{\Omega}} \sum_{y\in\Omega'} \Big|\mathrm{H}(\samp{\theta}(\cdot \mid \yy_{<t}))\\
    & \hspace{2.1cm} +\log \samp{\theta}(y \mid \yy_{<t})\Big| \nonumber\\
    &\text{where} \sum_{y \in \Omega'} \samp{\theta}(y \mid \yy_{<t}) \geq \tau. \nonumber
\end{align*} 
In other words, $\Omegat{typical}$ restricts samplings to tokens whose individual log-probabilities lie closest to the (sub)word-level entropy, with a combined probability mass of at least $\tau$, the parameter value.

\textbf{$\eta$-sampling}~\cite{hewitt-etal-2022-truncation} relies on: 
\begin{align}
    \Omegat{\eta} = \{y \in \bar{\Omega} \mid \samp{\theta}(y \mid \yy_{<t}) > \eta_t\},
\end{align}
where $\eta_t = \min 
\left(\epsilon, \sqrt{\epsilon}\exp(-\mathrm{H}\left(\samp{\theta}(\cdot \mid \yy_{<t})\right))\right)$. 
This corresponds to a subset of tokens with probability greater than $\eta_t$, a function of the local conditional entropy $\mathrm{H}$ of the current context parametrized by $\epsilon$.

\section{Measuring Diversity \label{sec:measuring diversity}}

\subsection{Evaluating Text Generation}
Standard evaluation metrics such as \emph{perplexity}-the exponentiated average negative log-likelihood of a sequence- primarily measure a model’s ability to predict each next token in isolation and thus capture only local, short-range dependencies. They overlook higher-level structural properties—such as syntactic richness or discourse coherence, that are crucial to our perception of naturalness. Our aim here is rather to assess if the produced outputs exhibit the same statistical tendencies as human samples. To this end, we follow the work done by \citet{meister-cotterell-2021-language} and measure the following metrics, aimed at capturing lexical diversity.
These are functions of the input text \yy, $N$ represents the total text length, and $V$ the set of unique words it contains.

\textbf{MTLD (Measure of Textual Lexical Diversity)}: Estimates the average number of tokens\footnote{Lexicographic metrics are defined based on the number of \emph{tokens}, which count word occurrences and \emph{types}, which correspond to unique word occurrences \citep{baayen-2000-words}.} needed before the Type–Token Ratio (TTR) falls below a threshold $\Theta$:
\begin{align*}
  \mathrm{MTLD} = \frac{1}{m}\sum_{j=1}^{m} n_j,
\end{align*}
where spans of length $n_j$ end when $\frac{\lvert\{\text{types in span }j\}\rvert}{n_j}<\Theta$. Our experiments use $\Theta=0.72$ \citep{mccarthy-etal-2010-MTLDVocdDHDD}.

\textbf{Hapax Legomena Ratio} measures the proportion of types that occur exactly once: 
\begin{align*}
  \mathrm{HLR} = \frac{V_1}{\rvert \Omega \lvert},
\end{align*}
with $V_1=\lvert\{w : \mathrm{freq}(w)=1\}\rvert$ and $\rvert\Omega \lvert$ is the vocabulary size.

\textbf{Simpson’s Diversity Index} \cite{simpson-etal-1949-MeasurementDiversity}: Gives the probability that two randomly chosen tokens (with replacement) are of the same type, the lower the value, the more diverse the text is.
\[
  D = \sum_{i=1}^{\rvert\Omega\lvert}{prop}_i^2,
\]
were $prop_i$ is the proportion of token $y_i$ in the text.

\textbf{Zipfian Exponent ($\alpha$)} \citep{zipf-etal-1950-HumanBehaviorPrinciple}: Quantifies how steeply token frequency $f_r$ decays with rank $r$ under Zipf’s law; larger $\alpha$ means faster drop-off:
\begin{align*}
  \log f_{(r)} &= -\alpha\,\log r + C,\\
  \alpha &= -\frac{\mathrm{Cov}(\log r,\log f_{(r)})}{\mathrm{Var}(\log r)}.
\end{align*}

\textbf{Heaps’ Law Exponent ($\beta$)} \citep{herdan-1960-type}: Models the rate at which new tokens appear; how distinct word count grows with the total tokens as $V(N)\approx K\,N^\beta$ with $K$ a constant:
\begin{align*}
  \log V_i &= \beta\,\log N_i + C,\\
  \beta &= \frac{\mathrm{Cov}(\log N_i,\log V_i)}{\mathrm{Var}(\log N_i)}.
\end{align*}
Where   $N_i$ is the total number of tokens in sample $i$, and 
  $V_i$ the number of distinct tokens in sample $i$.

\subsection{Comparing Probability Distributions \label{ssec:indicators}}

To better monitor the effect of each sampling adapters on the original probability distribution, we consider multiple measures of divergence between probability distributions $p$ and $q$. These metrics will also be used in the analysis of unsupervised detectors (\textsection{}~\ref{ssec:unsupervised}). 

\textbf{Total Variation Distance}: Measures the maximum discrepancy between probabilities assigned by the distributions (half the $L^1$ difference):
\[
  \mathrm{TV}(p,q) = \tfrac12 \sum_{\forall\,y} \lvert p(y)-q(y)\rvert.
\]

\textbf{Cross-Entropy}: Captures the average number of bits needed to encode samples from $p$ when using coding optimized for $q$:
\[
  CE(p,q) = -\sum_{\forall\,y} p(y)\,\log q(y).
\]

\textbf{Kullback-Leibler Divergence}: Computes the relative increase in entropy when using $q$ to approximate $p$: 
\[
  D_{\mathrm{KL}}(p\|q) = \sum_{\forall\, y} p(y)\,\log\frac{p(y)}{q(y)}.
\]
$\forall p, q, D_{\mathrm{KL}}(p,q) \ge 0$ and $D_{\mathrm{KL}}(p\|q) = 0 \Leftrightarrow p =q $ \citep{Cover91}.

\textbf{Rényi Divergence ($\alpha>0,\alpha\neq1$)}: Generalizes KL divergence (corresponding to $\alpha=1$), emphasizing different parts of the distribution depending on the value of $\alpha$. $\alpha>1$ emphasizes high‐probability (\qt{head}) events of $P$, while $\alpha<1$ boosts small $P(x)$, making the divergence sensitive to mismatches in the tail of the distribution:

\[
  D_{\alpha}(p\|q) = \frac{1}{\alpha-1}\,\log\Bigl(\sum_{\forall\, y} p(y)^\alpha\,q(y)^{1-\alpha}\Bigr).
\]

\textbf{Euclidean (L$_2$) Distance}: Considers $p$ and $q$ as vectors in $\mathbb{R}^{\lvert\Omega\lvert}$ and measures $L^2$ norm.
\[
  \lVert p-q\rVert_2 = \sqrt{\sum_{\forall\, y} \bigl(p(y)-q(y)\bigr)^2}.
\]

\section{Datasets \label{sec:datasets}}

\subsection{Human and Artificial Texts \label{ssec:textgen}}
In order to evaluate the effects of sampling on realistic data, generated with high-quality models, we randomly picked 2,000 English texts from the human samples of the RAID dataset, spanning across 11 different domains, along with their corresponding prompts. Then, using \llama \cite{llama3herdmodels}, we generated 37 artificial counterparts to each of these samples: 36 use the aforementioned sampling adapters, with hyperparameter values detailed in Table~\ref{tab:sampling-values} while the remaining one uses vanilla ancestral sampling. We split them into training and test set using a 1:1 ratio.

\begin{table}[ht]
\centering
\resizebox{\linewidth}{!}{
\begin{tabular}{l *{6}{c}}
\hline
\textbf{Sampling adapters} \\
\hline
Temperature        
  & 0.5   & 0.7   & 0.9   & 1.1   & 1.2   & 1.3   \\
Repetition penalty 
  & 1.05  & 1.10  & 1.15  & 1.20  & 1.25  & 1.30        \\
Top-\(k\) sampling          
  & 10    & 20    & 50    & 75    & 100   & 1000        \\
Top-\(p\) sampling         
  & 0.3   & 0.5   & 0.7   & 0.8   & 0.9   & 0.95        \\
Typical sampling         
  & 0.3   & 0.5   & 0.7   & 0.8   & 0.9   & 0.95        \\
\(\eta\) sampling  
  & $10^{-4}$ & $10^{-3}$ & $5\times10^{-3}$ 
  & 0.01 & 0.05 & 0.1        \\
\hline
\end{tabular}
}
\caption{Sampling adapters and tested parameter values, to which we added ancestral sampling}
\label{tab:sampling-values}
\end{table}

The prompts used are the ones carefully crafted by \citet{dugan-etal-2024-raid}; they are removed from the generated texts when saving the outputs. Generation was performed using Huggingface’s \verb|.generate()| function, where all these sampling adapters are implemented, with the value of \verb|max_tokens| set to 512 to approximately match the lengths of the human samples.

\subsection{Sampling Parameters Impact Lexical Diversity \label{ssec:dataset-diversity}}

\begin{table}[ht]
\centering
\resizebox{\linewidth}{!}{
\setlength{\tabcolsep}{3pt}
\begin{tabular}{@{}l|rrrrrrr@{}}
\toprule
Parameter                 &   MTLD & Hapax$^{\star}$ & Simpson$^\star$& Zipf\(_\alpha\) & Heaps$^\star$ & Avg.\ length & Perplexity \\
\midrule
\hr human                     &  94.60 &               34.90 &                 0.66 &         1.20 &              59.46 &    389.38  &        14.32 \\

temperature 0.50          &  28.24 &               30.56 &                 0.96 &         1.40 &              67.74 &       422.44 &      2.64 \\
temperature 0.70          &  38.50 &               32.56 &                 0.87 &         1.29 &              64.60 &       386.63 &      4.10 \\
\ok temperature 0.90      &  70.20 &               36.09 &                 0.73 &         1.19 &              64.75 &       378.32 &     11.07 \\
temperature 1.00          & 121.41 &               38.09 &                 0.57 &         1.12 &              66.50 &       375.54 &     38.08 \\
temperature 1.10          & 684.80 &               38.88 &                 0.22 &         1.02 &              67.42 &       384.26 &    612.48 \\
temperature 1.20          &7081.06 &               35.24 &                 0.04 &         0.95 &              64.37 &       411.83 &   7048.55 \\
temperature 1.30          &15264.92&               32.51 &                 0.02 &         0.91 &              63.55 &       443.49 &  14475.99 \\

repetition-penalty 1.05   & 428.22 &               38.19 &                 0.31 &         1.10 &              63.85 &       395.76 &    107.55 \\
repetition-penalty 1.10   &1600.29 &               36.28 &                 0.13 &         1.08 &              60.95 &       419.39 &    221.67 \\
repetition-penalty 1.15   &3020.03 &               34.92 &                 0.06 &         1.08 &              60.31 &       432.29 &    383.13 \\
repetition-penalty 1.20   &4373.57 &               33.23 &                 0.04 &         1.08 &              58.84 &       450.55 &    547.10 \\
repetition-penalty 1.25   &5098.23 &               32.26 &                 0.03 &         1.08 &              57.53 &       459.39 &    621.26 \\
repetition-penalty 1.30   &5570.62 &               31.84 &                 0.03 &         1.08 &              57.36 &       474.47 &    737.22 \\

top-k 10                  &  51.04 &               32.54 &                 0.79 &         1.27 &              64.86 &       357.78 &      4.98 \\
top-k 20                  &  63.60 &               33.23 &                 0.76 &         1.25 &              63.07 &       370.95 &      6.38 \\
top-k 50                  &  75.68 &               34.81 &                 0.71 &         1.23 &              62.59 &       370.98 &     16.04 \\
top-k 75                  &  76.64 &               35.41 &                 0.68 &         1.22 &              62.42 &       365.80 &      9.63 \\
\ok top-k 100             &  81.71 &               36.02 &                 0.68 &         1.23 &              63.72 &       371.80 &     11.12 \\
top-k 1000                & 102.50 &               36.04 &                 0.62 &         1.17 &              64.64 &       385.51 &     73.69 \\

top-p 0.30                &  24.00 &               31.15 &                 1.01 &         1.48 &              69.00 &       471.39 &      3.30 \\
top-p 0.50                &  28.75 &               30.79 &                 0.95 &         1.39 &              65.24 &       442.06 &      2.71 \\
top-p 0.70                &  36.60 &               33.62 &                 0.87 &         1.31 &              63.37 &       397.86 &      3.91 \\
top-p 0.80                &  51.07 &               33.74 &                 0.81 &         1.26 &              65.10 &       383.93 &      6.48 \\
top-p 0.90                &  69.83 &               36.17 &                 0.74 &         1.21 &              63.22 &       376.26 &     10.41 \\
\ok top-p 0.95            &  83.70 &               36.44 &                 0.67 &         1.19 &              64.58 &       372.32 &     19.07 \\

typical-sampling 0.30     &  74.78 &               35.92 &                 0.67 &         1.23 &              60.15 &       404.97 &     13.70 \\
typical-sampling 0.50     &  60.16 &               35.47 &                 0.75 &         1.23 &              63.69 &       392.03 &      8.87 \\
typical-sampling 0.70     &  52.76 &               34.45 &                 0.80 &         1.23 &              63.59 &       381.17 &      6.83 \\
typical-sampling 0.80     &  53.77 &               34.55 &                 0.81 &         1.24 &              62.24 &       387.65 &      6.49 \\
typical-sampling 0.90     &  65.92 &               36.05 &                 0.73 &         1.22 &              64.10 &       377.71 &    306.77 \\
\ok typical-sampling 0.95 &  87.52 &               36.51 &                 0.68 &         1.19 &              64.54 &       380.18 &     19.70 \\

eta-sampling 0.01         &  63.35 &               35.10 &                 0.77 &         1.24 &              63.39 &       378.02 &      9.40 \\
eta-sampling 0.05         &  47.96 &               33.93 &                 0.83 &         1.27 &              65.07 &       383.29 &      5.08 \\
eta-sampling 0.10         &  43.03 &               33.47 &                 0.84 &         1.28 &              63.25 &       385.43 &      4.51 \\
eta-sampling 1e-3         &  80.53 &               35.86 &                 0.70 &         1.21 &              65.34 &       380.43 &     10.87 \\
eta-sampling 5e-3         &  65.60 &               35.06 &                 0.75 &         1.23 &              63.68 &       372.26 &      9.91 \\
\ok eta-sampling 1e-4     &  93.45 &               36.53 &                 0.65 &         1.18 &              64.31 &       369.62 &     17.38 \\

\bottomrule
\end{tabular}
}
\caption{Lexical diversity within our dataset. $\star$ Hapax, Simpson, and Heaps’ values are multiplied by \(10^2\). Length is in \llama tokens. Highlighted in green are parameters that come close to the human standards (in yellow).}
\label{tab:diversity-metrics}
\end{table}

Table~\ref{tab:diversity-metrics} reports all 5 measures of diversity mentioned in Section~\ref{sec:measuring diversity} for every generated dataset, using the parameters mentioned in Table~\ref{tab:sampling-values}. We use the human samples as a gold‐standard profile of diversity. Relative to that baseline, we make the following observations:

\textbf{Temperature:} at $T=1.0$, MTLD and Simpson come reasonably close to human diversity. Cooler settings (0.5–0.7) under‐diversify, while hotter settings (1.1–1.3) overshoot dramatically, leading to non-sensical outputs.

\textbf{Repetition Penalty:} a mild penalty (1.05) drives MTLD to over $4$ times the average human span length, and halves Simpson, creating unnaturally long yet (ironically) repetitive runs. Stronger penalties stray even further from the human profile, often leading to nonsensical text.

\textbf{Truncation Sampling.} For \textbf{top-k}, as $k$ grows, the lexical distribution becomes more disparate, surpassing human standards as $k$ reaches $1,000$. \textbf{Top-p} and \textbf{Typical Sampling} behave similarly, with values closer to $1$ approaching the human reference. Regarding \textbf{$\eta$-sampling}, at $\eta=1\times10^{-4}$, MTLD and Simpson's index nearly match human levels and the Hapax ratio even slightly exceeds it — indicating very human‐like generation. Larger values lead to lower MTLD and higher Simpson, yielding more repetitions than in human samples.

In summary, $\eta=10^{-4}$, $T=1.0$, $k=100$, $p=0.95$, and $\tau=0.95$ yield diversity metrics closest to human text, whereas extreme repetition penalties or overly permissive settings push MTLD and perplexity far beyond—and Simpson far below— their natural (human) ranges. For further information, correlation between these measures and unsupervised detectors performance is reported in Appendix~\ref{app:correlation-details} Table~\ref{tab:pearson_diversity_correlations}.

\subsection{Impact on Human Evaluation} \label{ssec:human-eval}
Both extreme diversity regimes degrade quality in opposite ways. When diversity is too high, the sampler select tokens from very low-probability regions, yielding nonsensical outputs with grammatical errors and even language changes. Such texts are typically marked by depressed Simpson and inflated MTLD/Hapax. When diversity is too low, probability mass concentrates on frequent types, driving up Simpson and suppressing MTLD, which manifests as repetitive n-grams that feel artificial. By contrast, keeping these metrics within the human range produces fluent and coherent text; see Appendix~\ref{app:generation-examples} for examples.

\section{Experiments \label{sec:experiments}}

\begin{table*}[ht]
\centering
\resizebox{0.75\linewidth}{!}{
\setlength{\tabcolsep}{3pt}
\begin{tabular}{l|*{9}{c}}
                         & temp<1       & temp=1        & temp>1        & rep-penalty  & top-k         & top-p         & typical       & $\eta$        & mixture       \\ \hline
eta                      & \underline{0.95} & 0.87          & 0.64          & 0.72         & \underline{0.94} & 0.94          & \underline{0.94} & \underline{0.94} & 0.88          \\ 
rep-penalty              & 0.57          & 0.70          & \underline{0.98} & 0.96         & 0.60          & 0.57          & 0.59          & 0.59          & 0.69          \\ 
temp<1                   & \underline{0.95} & 0.79          & 0.63          & 0.68         & 0.91          & 0.94          & 0.91          & 0.92          & 0.85          \\ 
temp=1                   & 0.88          & \textbf{0.96}   & \textbf{0.99}   & \textbf{0.99}   & 0.92          & 0.87          & 0.90          & 0.90          & \underline{0.92} \\ 
temp>1                   & 0.52          & 0.61          & 0.95          & 0.88         & 0.54          & 0.52          & 0.53          & 0.53          & 0.62          \\ 
top-k                    & \underline{0.95} & 0.89          & 0.68          & 0.76         & \textbf{0.95}   & \underline{0.95} & \underline{0.94} & \textbf{0.95}   & 0.89          \\ 
top-p                    & 0.93          & 0.77          & 0.61          & 0.65         & 0.89          & 0.93          & 0.89          & 0.90          & 0.83          \\ 
typical                  & \underline{0.95} & 0.91          & 0.73          & 0.82         & \underline{0.94} & 0.94          & \underline{0.94} & \underline{0.94} & 0.90          \\ 
mixture                  & \textbf{0.96}   & \underline{0.95} & \underline{0.98} & \underline{0.98} & \textbf{0.95}   & \textbf{0.96}   & \textbf{0.95}   & \textbf{0.95}   & \textbf{0.96}   \\
\end{tabular}
}
\caption{Aggregated sampling‐parameter accuracy values. RoBERTa is trained with texts generated with the “row” parameter value, and tested with the “column” one. Bolded is \textbf{Best} value per column, underlined is \underline{second best}.}
\label{tab:agg-roberta-accuracy}
\end{table*}

\subsection{Artificial text detectors}

We evaluate three identification methods.\footnote{We also considered DNA-GPT \citep{yang-etal-2024-dnagpt}, a rewriting approach, but the results were poor, see Appendix~\ref{app:dna-gpt}.} As most supervised approaches involve training a RoBERTa-based \citep{liu-etal-2019-roberta} classifier, we adopted the same strategy to assess this setup. We fine-tuned a RoBERTa-base model using \verb|transformers.Trainer|\footnote{\url{https://huggingface.co/docs/transformers/main_classes/trainer}} along with the default \verb|TrainingArguments| : a learning rate of 5e-05 for 3 epochs, using AdamW as the optimizer.
In the unsupervised setting, we tested both \bino \citep{hans-etal-2024-SpottingLLMsBinoculars} and FastDetectGPT \citep{bao-etal-2024-fastdetectgpt}, as they are the top performers reported in the RAID leaderboard \citep{dugan-etal-2024-raid}. 

In all these settings, we place ourselves in a \qt{best-case scenario}, as the supervised case involves training and testing on the same generator model, and we use \llama ($q$) and \llama-Instruct ($r$) as the underlying models for both \bino{} and FastDetectGPT, per their author’s suggestion. \citet{bao-etal-2024-fastdetectgpt} recommend detecting with the same model used for generation, and \citet{hans-etal-2024-SpottingLLMsBinoculars} advise to use a model and its Instruct version as an ideal combination.

\subsection{Supervised Detection \label{ssec:supervised}}
\begin{figure}[ht]
    \centering
    \includegraphics[width=\linewidth]{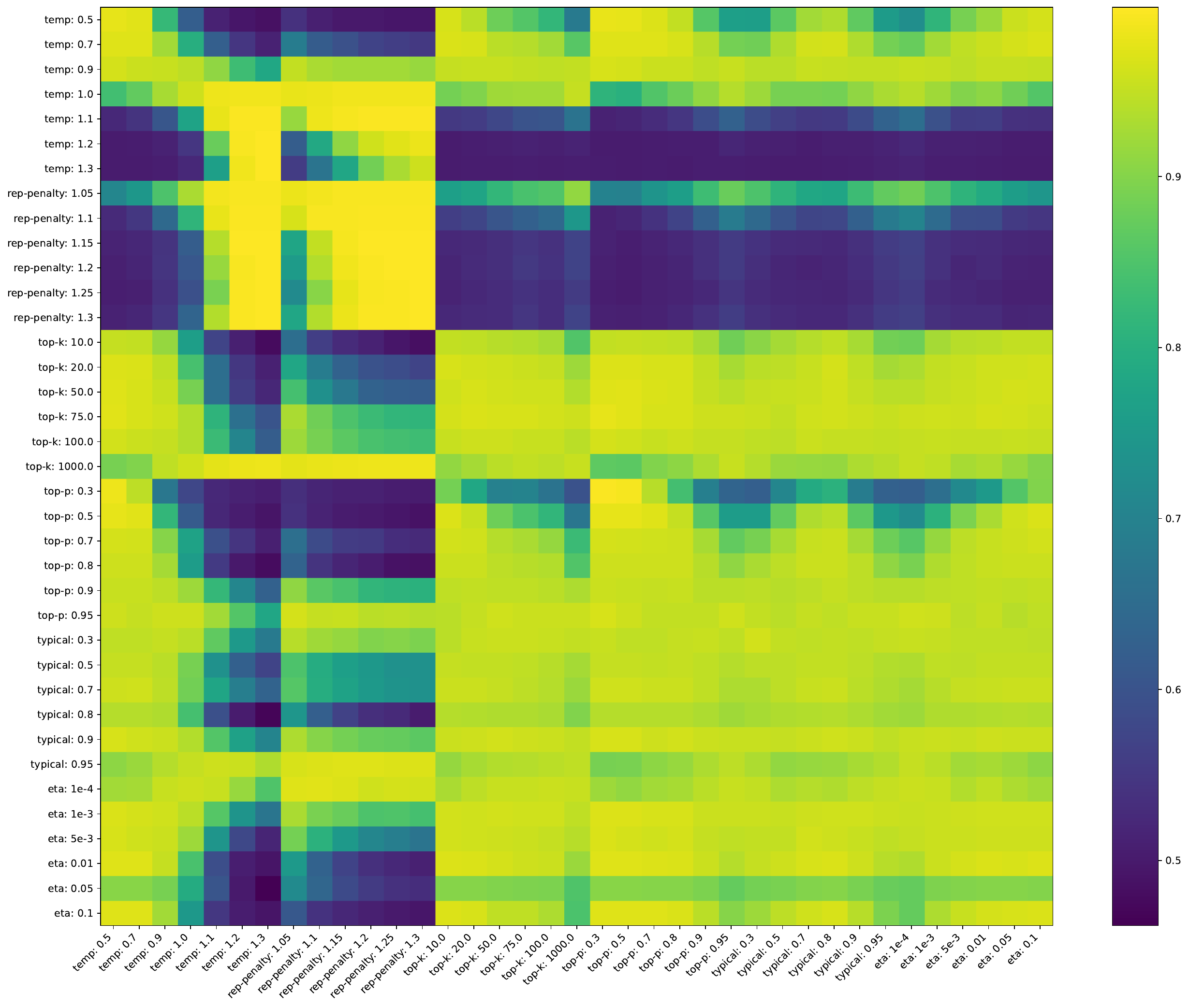}
    \caption{Heatmap of accuracy detection rates when training using data generated with the row parameter and testing on the column one. For further details, this heatmap is displayed in full in Appendix~\ref{app:supervised-details} Figure~\ref{fig:roberta-accuracy-details}.}
    \label{fig:roberta-accuracy}
\end{figure}

Figure~\ref{fig:roberta-accuracy} displays a heatmap representing what happens when training a RoBERTa classifier to distinguish between human and generated texts using one parameter (like temperature=$T$), then testing if it can identify machine outputs generated using another (like typical=$\rho$).
It appears that detectors trained on repetition-penalty and high temperatures do not generalize well (and vice-versa), whereas parameters that are often recommended for their \qt{better} generation quality (values of $p$, $T$ and typical close to $1$, small values of $\eta$ and high values of $k$) lead to well-rounded performance.

Training on a uniformly sampled mixture of the training data seems to solve misclassification issues almost entirely, yielding an accuracy of at least 95\% for all tested parameters. See Table~\ref{tab:agg-roberta-accuracy} and Appendix Figure~\ref{fig:roberta-accuracy-details} for further details.

Training on a mixture of parameters that are close to human levels (the rows highlighted in green Table~\ref{tab:diversity-metrics}) lead to a supervised detector with good all-around performance, yielding an accuracy of 94\% averaged on all parameters. Nonsensical generations due high temperatures are bringing down the mean performance with accuracies of 87\% and 80\% for T=1.2 and T=1.3 respectively. 

A small follow-up experiment however reveals that training with a mixture of data still yields to a very brittle detector. For this experiment, we changed the \emph{human} subset of the RAID test data (comprising BBC News articles \citep{bbc-news}) and replaced it with News articles from the CCNews subset of \citet{common-crawl-news}. We observed a strong decrease in detection accuracy, which drops down to 72\% on average. This introduces a small domain mismatch, as only about a tenth of the training texts were part of a news dataset, and suggests that our supervised classifier is mostly overfitting its human training dataset, and struggles with other sources of texts, even from the same domain. Since the goal of this study is about sampling parameters, we will not dig further down this path.

\subsection{Unsupervised Detection \label{ssec:unsupervised}}

\begin{table*}[ht!]
  \centering
  \captionsetup[subtable]{font=small,labelformat=empty}
  \setlength{\tabcolsep}{3pt}

  \begin{subtable}[t]{0.5\linewidth}
    \centering\scriptsize
    \begin{tabular}{l c c G c c c}
      \toprule
      & \multicolumn{6}{c}{Temperature} \\
      \cmidrule(l){2-7}
      Method             & 0.5    & 0.7    & 0.9    & 1.1    & 1.2    & 1.3    \\
      \midrule
      Fast               & 0.9468 & 0.9419 & 0.8729 & 0.3693 & 0.2608 & 0.1672 \\
      \bino         & 0.9928 & 0.9918 & 0.9449 & 0.0200 & 0.0056 & 0.0019 \\
      Fast-uniform       & 0.9713 & 0.9649 & 0.9029      & 0.3529 & 0.2098 & 0.1045 \\
      \bino-uniform & 0.9937 & 0.9927 & 0.9571 & 0.0342 & 0.0101 & 0.0036 \\
      \bottomrule
    \end{tabular}
  \end{subtable}%
  \hfill
  \begin{subtable}[t]{0.5\linewidth}
    \centering\scriptsize
    \begin{tabular}{l*{6}{c}}
      \toprule
      & \multicolumn{6}{c}{Repetition penalty} \\
      \cmidrule(l){2-7}
      Method             & 1.05   & 1.10   & 1.15   & 1.20   & 1.25   & 1.30   \\
      \midrule
      Fast               & 0.4443 & 0.3235 & 0.3897 & 0.4774 & 0.5171 & 0.5699 \\
      \bino         & 0.0711 & 0.0182 & 0.0157 & 0.0157 & 0.0164 & 0.0182 \\
      Fast-uniform       & 0.4189 & 0.2979 & 0.3678 & 0.4552 & 0.4967 & 0.5473 \\
      \bino-uniform & 0.0986 & 0.0319 & 0.0300 & 0.0326 & 0.0342 & 0.0380 \\
      \bottomrule
    \end{tabular}
  \end{subtable}

  \begin{subtable}[t]{0.5\linewidth}
    \centering\scriptsize
    \begin{tabular}{l c c c c G c}
      \toprule
      & \multicolumn{6}{c}{Top-\(k\)} \\
      \cmidrule(l){2-7}
      Method             & 10     & 20     & 50     & 75     & 100    & 1000   \\
      \midrule
      Fast               & 0.9198 & 0.9038 & 0.8659 & 0.8280 & 0.8421 & 0.7555 \\
      \bino         & 0.9916 & 0.9834 & 0.9544 & 0.9214 & 0.9188 & 0.6355 \\
      Fast-uniform       & 0.9483 & 0.9321 & 0.8946      & 0.8598 & 0.8719 & 0.7686 \\
      \bino-uniform & 0.9922 & 0.9846 & 0.9596     & 0.9306 & 0.9310 & 0.6800 \\
      \bottomrule
    \end{tabular}

  \end{subtable}%
  \hfill
  \begin{subtable}[t]{0.5\linewidth}
    \centering\scriptsize
    \begin{tabular}{l c c c c c G}
      \toprule
      & \multicolumn{6}{c}{Top-\(p\)} \\
      \cmidrule(l){2-7}
      Method             & 0.3    & 0.5    & 0.7    & 0.8    & 0.9    & 0.95   \\
      \midrule
      Fast               & 0.9533 & 0.9680 & 0.9604 & 0.9395 & 0.8898 & 0.8254 \\
      \bino         & 0.9942 & 0.9956 & 0.9975 & 0.9928 & 0.9726 & 0.8625 \\
      Fast-uniform       & 0.9751 & 0.9829 & 0.9779 & 0.9616 & 0.9190      & 0.8526 \\
      \bino-uniform & 0.9941 & 0.9959 & 0.9978 & 0.9938 & 0.9784      & 0.8868 \\
      \bottomrule
    \end{tabular}
  \end{subtable}

  \begin{subtable}[t]{0.5\linewidth}
    \centering\scriptsize
    \begin{tabular}{l c c c c c G}
      \toprule
      & \multicolumn{6}{c}{Typical sampling} \\
      \cmidrule(l){2-7}
      Method             & 0.3    & 0.5    & 0.7    & 0.8    & 0.9    & 0.95   \\
      \midrule
      Fast               & 0.8776 & 0.9308 & 0.9329 & 0.9382 & 0.8874 & 0.8362 \\
      \bino         & 0.9047 & 0.9851 & 0.9887 & 0.9933 & 0.9630 & 0.8703 \\
      Fast-uniform       & 0.9050 & 0.9544      & 0.9557 & 0.9609 & 0.9144 & 0.8606      \\
      \bino-uniform & 0.9375 & 0.9904 & 0.9903 & 0.9950 & 0.9683 & 0.8924 \\
      \bottomrule
    \end{tabular}
  \end{subtable}%
  \hfill
  \begin{subtable}[t]{0.5\linewidth}
    \centering\scriptsize
    \begin{tabular}{l G c c c c c}
      \toprule
      & \multicolumn{6}{c}{\(\eta\)-sampling} \\
      \cmidrule(l){2-7}
      Method             & 1e-4  & 1e-3  & 5e-3  & 0.01  & 0.05  & 0.10  \\
      \midrule
      Fast               & 0.7799 & 0.8553 & 0.9019 & 0.9102 & 0.9466 & 0.9487 \\
      \bino         & 0.7672 & 0.9355 & 0.9779 & 0.9839 & 0.9965 & 0.9928 \\
      Fast-uniform       & 0.8006 & 0.8827      & 0.9288 & 0.9348 & 0.9676 & 0.9682 \\
      \bino-uniform & 0.7958 & 0.9454      & 0.9816 & 0.9855 & 0.9970 & 0.9937 \\
      \bottomrule
    \end{tabular}
  \end{subtable}

  \caption{ROC AUC of \bino{} and Fast(DetectGPT) across decoding parameters, by adapter family. Highlighted in yellow are the settings the closest to the human diversity values (see Table~\ref{tab:diversity-metrics}).}
  \label{tab:auroc-subtables}
\end{table*}

The unsupervised detection methods considered here involve the combination of \emph{two detector models} $q$ and $r$.  For these methods, the detection score averages over tokens the (log)-difference of the surprisal of the current token computed by the \emph{main model} $q$ with the cross-entropy between the \emph{auxiliary model }$r$ and $q$. Formally, for input sequence $\mathbf{y} =  \langle y_0, y_1, \ldots y_T \rangle$,
the \bino{} score $B_{q,r}(\mathbf{y})$ is defined by : 
\begin{equation}
    B_{q,r}(\mathbf{y}) \triangleq   \frac{ \sum_{t=1}^{T}  \sum_{y\in\Omega} \mathbbm{1} [y=y_t]
  \mathcal{L}_q(y_t |\mathbf{y}_{<t}) }{ 
   \sum_{t=1}^{T} 
    \sum_{y\in\bar{\Omega}} \displaystyle  r (y | \mathbf{y}_{<t})\mathcal{L}_q(y | \mathbf{y}_{<t})}
    \label{eq:binoculars-score}
\end{equation}
with $\mathcal{L}_q ( y_t | \mathbf{y}_{<t}) = - \log q( y_t | \mathbf{y}_{<t})$, and $q (y | \mathbf{y}_{<t})$ and $r (y | \mathbf{y}_{<t})$ representing the probabilities assigned by models $q$ and $r$, respectively, to token $y$ conditioned on the current context $\mathbf{y}_{<t}$. The summation over $\bar{\Omega}$ in Eq.~\eqref{eq:binoculars-score} implies that this score is only valid when $q$ and $r$ have the same support, meaning that they share a common underlying vocabulary and tokenizer.

FastDetectGPT’s scoring method is using a similar idea but normalizes scores and approximates the cross-entropy, more details are in Appendix~\ref{app:fast-equation}.
Unlike in the supervised case, these methods do not directly output a label but rather a score, which is why we report AUROC as our metric for these methods (computed with \verb|scikit-learn| \cite{pedregosa-2011-scikit-learn}). These scores cannot be compared to the accuracy values reported in Section~\ref{ssec:supervised}. AUROC values close to $0$ and $1$ indicate that the scores histograms are well separated, and scores close to $0.5$ indicate that the human and machine texts are indistinguishable from one another. 

\begin{figure}[ht]
  \centering
  \begin{subfigure}[b]{0.48\textwidth}
    \centering
    \includegraphics[width=0.48\textwidth]{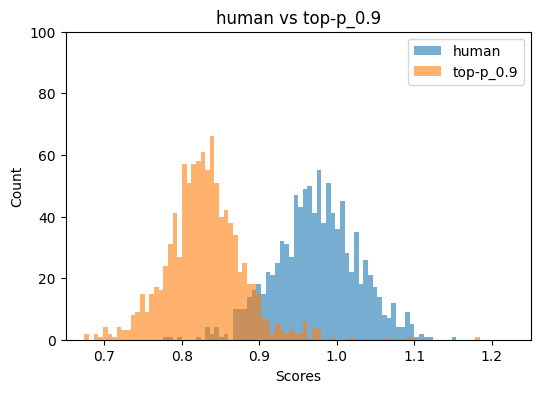}
    \hfill
    \includegraphics[width=0.48\textwidth]{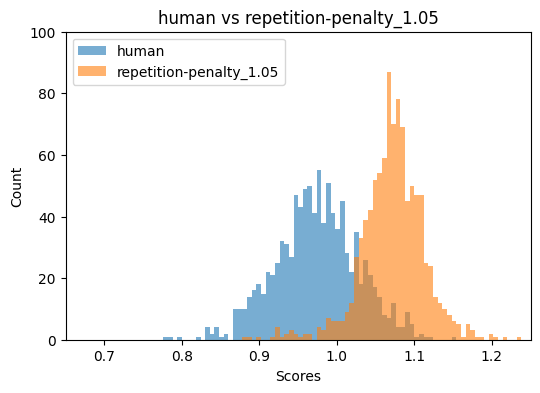}
    \caption{\bino{} scores.}
    \label{fig:bino-hist}
  \end{subfigure}
  \hfill
  \begin{subfigure}[b]{0.48\textwidth}
    \centering
    \includegraphics[width=0.48\textwidth]{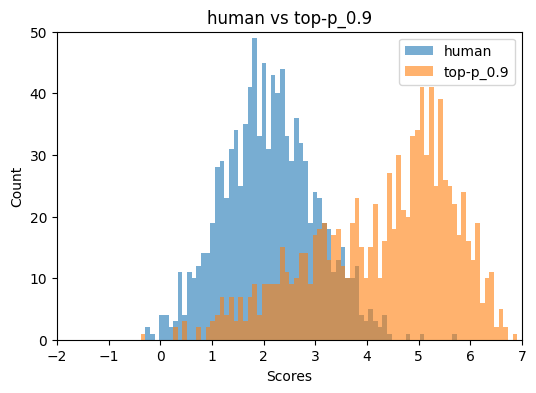}
    \hfill
    \includegraphics[width=0.48\textwidth]{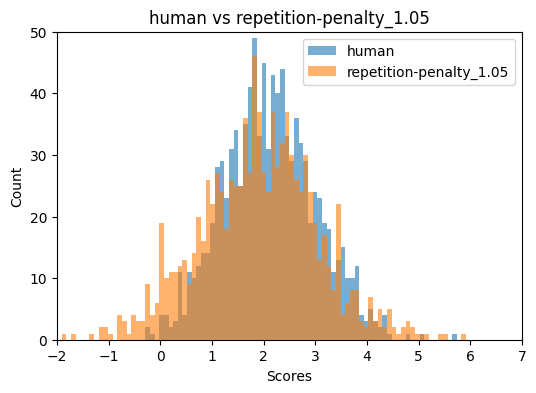}
    \caption{FastDetectGPT scores.}
    \label{fig:fast-hist}
  \end{subfigure}

  \caption{Scores histograms for top-$p=0.9$ (left) and repetition penalty $=1.05$ (right). For top-$p$, both detectors behave as intended, clearly separating human and machine scores. On the right, Binoculars switches the scores, going over the threshold value that was around 0.9 (this can be fixed by introducing a second theshold for high scores). For FastDetectGPT, the scores histograms are completely blended together, showing that repetition-penalty fully breaks that method.}
  \label{fig:scores-histograms}
\end{figure}

Table~\ref{tab:auroc-subtables} reports the performance of both methods on all texts generated with the sampling adapters of section~\ref{ssec:decoding}. Similar to the supervised setting, repetition penalty and temperatures above $1$ are troublemakers, making FastDetectGPT unable to separate human and machine-written samples, whereas the latter tend to become \qt{too human-like}, i.e., surprising, for \bino. An example is shown in Figure~\ref{fig:scores-histograms}. In most other settings, \bino{} outperforms FastDetectGPT by a very small margin. We hypothesize that these differences are due to the cross-entropy term being fully computed in Eq.~\ref{eq:binoculars-score}, allowing \bino{} to better deal with words that are in the tail of the distribution, as opposed to the empirical cross-entropy used in Eq.~\ref{eq:fast-score}.

However, fixing the issue with mixture models did not prove as successful as in the supervised case. For both scoring methods (Eq.~\ref{eq:binoculars-score} and Eq.~\ref{eq:fast-score}), we replaced the main model $q$  by $\tilde{q}$, a uniform mixture of all decoding-induced distributions considered in our study: $\tilde{q}\incontext = \frac{1}{N} \sum_1^N q_{i} \incontext$ where $q_{1},\ldots ,q_{N}$ = $q^{\eta = 1e-4}, \ldots ,q^{top-p=0.95}$, for all values of Table~\ref{tab:sampling-values}, plus ancestral sampling, for a grand total of 37 sampling settings.

The results of this mixture detector are reported in Table~\ref{tab:auroc-subtables}, on rows \qt{Fast-uniform} and \qt{\bino-uniform}. We observe that this modification of the main detector model improved the performance for both systems for all settings (albeit by a very small amount), with exception of the two troublesome ones (repetition penalty and temperature greater than 1).

\subsection{A weakness of unsupervised scores}

Results of Table~\ref{tab:diversity-metrics} show that the undetected generators $p$ are the more diverse ones. Yet, the whole appeal of two-model methods is their ability to deal with this diversity (e.g., for instance, the discussion in \citet[Sec. 3.2]{hans-etal-2024-SpottingLLMsBinoculars}).
To better grasp exactly which aspect of the adapted texts throws detectors into disarray, we correlate \bino{} and FastDetectGPT detection performance with two families of indicators introduced in \textsection{}~\ref{ssec:indicators}: the first compares the (adapted) generator $p$ and the main detector model $q$, using either samples from $p$ (i.e, some generated text), or the full distribution; the second focuses on the differences between the two detector models $q$ and $r$, with contexts sampled from $p$.\footnote{To avoid issues with support mismatch, we use an $\epsilon$-smoothed version of the probability distributions when necessary: $(p_{\epsilon} = \frac{p + \epsilon}{1 + \rvert\bar{\Omega}\lvert\cdot\epsilon})$. This trick is commonly used to ensure that the supports of the compared distributions match \citep{peters-etal-2019-sparse,martins-etal-2020-sparse,meister-etal-2023-efficacy}.} All indicators are averaged over tokens.


\begin{table}[ht]
\centering
\setlength{\tabcolsep}{3pt}
\small
\begin{tabular}{lcc}
\toprule
& \textbf{\bino} & \textbf{FastDetectGPT} \\
\midrule
\multicolumn{3}{l}{\textbf{Compares $q$ and samples from $p$}} \\
Perplexity                & $-0.458$ & $-0.611$ \\
Entropy                   & $-0.910$ & $-0.935$ \\
\midrule
\multicolumn{3}{l}{\textbf{Compares $q$ and $p$}} \\
TV                        & $-0.056$ & $+0.050$ \\
L$_2$      (Euclidean)    & $-0.254$ & $-0.118$ \\
Cross-entropy             & $+0.502$ & $+0.526$ \\
\midrule\midrule
\multicolumn{3}{l}{\textbf{Compares $q$ and $r$}} \\
KL (between models)       & $-0.850$ & $-0.898$ \\
Rényi ($\alpha=0.2$)       & $-0.890$ & $-0.921$ \\
Rényi ($\alpha=1.2$)       & $-0.828$ & $-0.882$ \\
\bottomrule
\end{tabular}
\caption{Pearson correlation coefficients between indicators and the AUROC of \bino{} and FastDetectGPT.}
\label{tab:pearson_correlations}
\end{table}

From Table~\ref{tab:pearson_correlations}, one can see that perplexity is not the best predictor of detection performance, going in the way \citet{hans-etal-2024-SpottingLLMsBinoculars} intended. The distance between the model’s original distribution ($q$) and its adapted version ($p$) is not better correlated: extreme truncations induce large divergence values, yet are very easy to detect. The two most important factors seem to be the (averaged) entropy of $q$ over the text and the divergence between the two detector models: when it is high, detection performance drops. This is because large divergences between these two models are interpreted as signals of a human text; it turns out that using the ``troublesome adaptors'' also yields large divergences for the value of $q$ and $r$ considered here - making artificial texts look like natural generations to these detection methods.

\section{Key Takeaways\label{sec:takeaways}}
\textit{Observation $N^{o}1$}. The detection of machine-written texts is not fully solved, even in supervised settings. As shown by the last experiment in \textsection~\ref{ssec:supervised}, human samples that are \qt{out-of-domain} (news articles from another source than the training data) are poorly identified by a strong RoBERTa classifier. In addition, it seems that some generation strategies, notably using repetition-penalties, need to be paid special attention, when preparing the training data for the detector.

\textit{Observation $N^{o}2$}. Existing benchmarks do not sufficiently represent all decoding strategies. If changing one parameter value (e.g., the temperature) seriously derails detectors, it is likely that they would also struggle to identify  more elaborate generation techniques such as Minimum Bayes Risk (MBR) decoding \cite{bertsch-etal-2023-mbr} or Monte Carlo Tree Search (MCTS) \cite{chaffin-etal-2022-ppl}.

\textit{Observation $N^{o}3$}. Two state-of-the-art unsupervised detection systems need their detector models to assign very similar probabilities to artificial texts; failures to do so will cause on overdetection of natural texts (Table~\ref{tab:pearson_correlations}). 
This explains why, for both systems, optimal detection scores are obtained with models from the same family (e.g., a pretrained model and the matching instruct version).


\section{Conclusion \label{sec:conclusion}}
In this work, we have presented a systematic study of how sampling-based decoding strategies affect the detectability of texts produced by large language models. Considering 37 different configurations—varying multiple generation parameters, 
we have shown that even modest adjustments in the (sub)word-level generation distribution can undermine state-of-the-art detectors, observing AUROC drops from near‐perfect down to null simply by tuning the repetition-penalty.

Our analysis sheds light on the underlying failure mechanisms: different sampling techniques induce distinctive distributional signatures (e.g., changes in token entropy or frequency tails) that current detectors are not able to recognize. This suggests that robustness in detection is still an open issue, both for supervised and unsupervised methods.

By exposing blind spots in current systems, we aim to steer the field toward more reliable and trustworthy methods for distinguishing human and machine-generated text, an ability that is bound to become essential in an era of fluent and accessible generative models.


\section*{Acknowledgements}
We’d like to thank the reviewers for their feedback and especially reviewer xqub for his insightful comments, helping us improve and refine unclear parts of our paper. ChatGPT was used to write comments and a README in the released code, in order to make sure it is understandable and re-usable by everyone.

This work was performed using HPC resources from GENCI–IDRIS (Grant 2023-AD011014903R1).

\section*{Limitations}
In this empirical study of the impact of the generation parameters and strategies used in text generation, we had to limit ourselves to a small number of parameters, and to a restricted set of values. Likewise, our study only considered one supervised architectures, and two unsupervised models. Finally, only text generation in English was considered, using \llama. These choices were made for the sake of limiting the computational expenses to a reasonable budget. We still believe that they are representative of the way language models are currently used to generate texts, and that our main conclusions --that text detection benchmarks should be expended to better assess detectors robustness-- would carry over more generally to other languages, models, and architectures.

\section*{Ethical Statement}

With the increasing availability of sophisticated text generation tools, the need for effective text detection systems is becoming ever more pressing, notably to combat fraud, plagiarism, deception and fake news on the Internet. Yet, artificial text detection tools are not infallible --as we have ourselves amply documented--  and consequently should not be used as the only basis for punitive actions or decisions that could affect individuals, notably students, or organizations. Detection methods must then always be complemented by human oversight and verification before taking any drastic measure, to ensure fairness of treatment.


\bibliography{anthology,custom}

\begin{thebibliography}{64}
\providecommand{\natexlab}[1]{#1}

\bibitem[{Abassy et~al.(2024)Abassy, Elozeiri, Aziz, Ta, Tomar, Adhikari, Ahmed, Wang, Mohammed~Afzal, Xie, Mansurov, Artemova, Mikhailov, Xing, Geng, Iqbal, Mujahid, Mahmoud, Tsvigun, Aji, Shelmanov, Habash, Gurevych, and Nakov}]{abassy-etal-2024-llm}
Mervat Abassy, Kareem Elozeiri, Alexander Aziz, Minh~Ngoc Ta, Raj~Vardhan Tomar, Bimarsha Adhikari, Saad El~Dine Ahmed, Yuxia Wang, Osama Mohammed~Afzal, Zhuohan Xie, Jonibek Mansurov, Ekaterina Artemova, Vladislav Mikhailov, Rui Xing, Jiahui Geng, Hasan Iqbal, Zain~Muhammad Mujahid, Tarek Mahmoud, Akim Tsvigun, and 5 others. 2024.
\newblock \href {https://doi.org/10.18653/v1/2024.emnlp-demo.35} {{LLM}-{D}etect{AI}ve: a tool for fine-grained machine-generated text detection}.
\newblock In \emph{Proceedings of the 2024 Conference on Empirical Methods in Natural Language Processing: System Demonstrations}, pages 336--343, Miami, Florida, USA. Association for Computational Linguistics.

\bibitem[{Antoun et~al.(2024)Antoun, Sagot, and Seddah}]{antoun-etal-2024-text}
Wissam Antoun, Beno{\^i}t Sagot, and Djam{\'e} Seddah. 2024.
\newblock \href {https://aclanthology.org/2024.lrec-main.665/} {From text to source: Results in detecting large language model-generated content}.
\newblock In \emph{Proceedings of the 2024 Joint International Conference on Computational Linguistics, Language Resources and Evaluation (LREC-COLING 2024)}, pages 7531--7543, Torino, Italia. ELRA and ICCL.

\bibitem[{Baayen(2001)}]{baayen-2000-words}
R.~Harald Baayen. 2001.
\newblock \href {https://doi.org/10.1007/978-94-010-0844-0} {\emph{Word {{Frequency Distributions}}}}, volume~18 of \emph{Text, {{Speech}} and {{Language Technology}}}.
\newblock Springer Netherlands, Dordrecht.

\bibitem[{Bao et~al.(2025)Bao, Zhao, He, and Zhang}]{bao-etal-2025-glimpse}
Guangsheng Bao, Yanbin Zhao, Juncai He, and Yue Zhang. 2025.
\newblock \href {https://openreview.net/forum?id=an3fugFA23} {Glimpse: Enabling white-box methods to use proprietary models for zero-shot {LLM}-generated text detection}.
\newblock In \emph{The Thirteenth International Conference on Learning Representations}.

\bibitem[{Bao et~al.(2024)Bao, Zhao, Teng, Yang, and Zhang}]{bao-etal-2024-fastdetectgpt}
Guangsheng Bao, Yanbin Zhao, Zhiyang Teng, Linyi Yang, and Yue Zhang. 2024.
\newblock \href {https://openreview.net/forum?id=Bpcgcr8E8Z} {Fast-detect{GPT}: Efficient zero-shot detection of machine-generated text via conditional probability curvature}.
\newblock In \emph{The Twelfth International Conference on Learning Representations}.

\bibitem[{Bertsch et~al.(2023)Bertsch, Xie, Neubig, and Gormley}]{bertsch-etal-2023-mbr}
Amanda Bertsch, Alex Xie, Graham Neubig, and Matthew Gormley. 2023.
\newblock \href {https://doi.org/10.18653/v1/2023.bigpicture-1.9} {It`s {MBR} all the way down: Modern generation techniques through the lens of minimum {B}ayes risk}.
\newblock In \emph{Proceedings of the Big Picture Workshop}, pages 108--122, Singapore. Association for Computational Linguistics.

\bibitem[{Bevendorff et~al.(2024)Bevendorff, Wiegmann, Karlgren, D{\"u}rlich, Gogoulou, Talman, Stamatatos, Potthast, and Stein}]{bevendorff-etal-2024-authorship}
Janek Bevendorff, Matti Wiegmann, Jussi Karlgren, Luise D{\"u}rlich, Evangelia Gogoulou, Aarne Talman, Efstathios Stamatatos, Martin Potthast, and Benno Stein. 2024.
\newblock {Overview of the ``Voight-Kampff'' Generative AI Authorship Verification Task at PAN and ELOQUENT 2024}.
\newblock In \emph{Working Notes of CLEF 2024 - Conference and Labs of the Evaluation Forum}, CEUR Workshop Proceedings. CEUR-WS.org.

\bibitem[{Chaffin et~al.(2022)Chaffin, Claveau, and Kijak}]{chaffin-etal-2022-ppl}
Antoine Chaffin, Vincent Claveau, and Ewa Kijak. 2022.
\newblock \href {https://doi.org/10.18653/v1/2022.naacl-main.215} {{PPL-MCTS}: {C}onstrained textual generation through discriminator-guided {MCTS} decoding}.
\newblock In \emph{Proceedings of the 2022 Conference of the North American Chapter of the Association for Computational Linguistics: Human Language Technologies}, pages 2953--2967, Seattle, United States. Association for Computational Linguistics.

\bibitem[{Clark et~al.(2021)Clark, August, Serrano, Haduong, Gururangan, and Smith}]{clark-etal-2021-thats}
Elizabeth Clark, Tal August, Sofia Serrano, Nikita Haduong, Suchin Gururangan, and Noah~A. Smith. 2021.
\newblock \href {https://doi.org/10.18653/v1/2021.acl-long.565} {All that`s {\textquoteleft}human' is not gold: Evaluating human evaluation of generated text}.
\newblock In \emph{Proceedings of the 59th Annual Meeting of the Association for Computational Linguistics and the 11th International Joint Conference on Natural Language Processing (Volume 1: Long Papers)}, pages 7282--7296, Online. Association for Computational Linguistics.

\bibitem[{{Common Crawl}(2016)}]{common-crawl-news}
{Common Crawl}. 2016.
\newblock News dataset available.
\newblock \url{https://commoncrawl.org/blog/news-dataset-available}.
\newblock [Online; accessed 19 May 2025].

\bibitem[{Cover and Thomas(2006)}]{Cover91}
T.~M. Cover and J.~A. Thomas. 2006.
\newblock \emph{Elements of Information Theory}, 2nd edition.
\newblock Wiley, New York, NY.

\bibitem[{Crothers et~al.(2023)Crothers, Japkowicz, and Viktor}]{crothers-etal-2023-machine}
Evan~N. Crothers, Nathalie Japkowicz, and Herna~L. Viktor. 2023.
\newblock \href {https://doi.org/10.1109/ACCESS.2023.3294090} {Machine-generated text: A comprehensive survey of threat models and detection methods}.
\newblock \emph{IEEE Access}, 11:70977--71002.

\bibitem[{Dubois et~al.(2025)Dubois, Yvon, and Piantanida}]{dubois-etal-2025-mosaicmultipleobserversspotting}
Matthieu Dubois, Fran{\c{c}}ois Yvon, and Pablo Piantanida. 2025.
\newblock \href {https://doi.org/10.18653/v1/2025.findings-acl.1244} {{MOSAIC}: Multiple observers spotting {AI} content}.
\newblock In \emph{Findings of the Association for Computational Linguistics: ACL 2025}, pages 24230--24247, Vienna, Austria. Association for Computational Linguistics.

\bibitem[{Dugan et~al.(2024)Dugan, Hwang, Trhl{\'i}k, Zhu, Ludan, Xu, Ippolito, and Callison-Burch}]{dugan-etal-2024-raid}
Liam Dugan, Alyssa Hwang, Filip Trhl{\'i}k, Andrew Zhu, Josh~Magnus Ludan, Hainiu Xu, Daphne Ippolito, and Chris Callison-Burch. 2024.
\newblock \href {https://doi.org/10.18653/v1/2024.acl-long.674} {{RAID}: A shared benchmark for robust evaluation of machine-generated text detectors}.
\newblock In \emph{Proceedings of the 62nd Annual Meeting of the Association for Computational Linguistics (Volume 1: Long Papers)}, pages 12463--12492, Bangkok, Thailand. Association for Computational Linguistics.

\bibitem[{Dugan et~al.(2023)Dugan, Ippolito, Kirubarajan, Shi, and Callison-Burch}]{dugan-etal-2023-roft}
Liam Dugan, Daphne Ippolito, Arun Kirubarajan, Sherry Shi, and Chris Callison-Burch. 2023.
\newblock \href {https://doi.org/10.1609/aaai.v37i11.26501} {Real or fake text?: Investigating human ability to detect boundaries between human-written and machine-generated text}.
\newblock \emph{Proceedings of the AAAI Conference on Artificial Intelligence}, 37(11):12763--12771.

\bibitem[{Dugan et~al.(2025)Dugan, Zhu, Alam, Nakov, Apidianaki, and Callison-Burch}]{dugan-etal-2025-genai}
Liam Dugan, Andrew Zhu, Firoj Alam, Preslav Nakov, Marianna Apidianaki, and Chris Callison-Burch. 2025.
\newblock \href {https://aclanthology.org/2025.genaidetect-1.45/} {{G}en{AI} content detection task 3: Cross-domain machine generated text detection challenge}.
\newblock In \emph{Proceedings of the 1stWorkshop on GenAI Content Detection (GenAIDetect)}, pages 377--388, Abu Dhabi, UAE. International Conference on Computational Linguistics.

\bibitem[{Fagni et~al.(2021)Fagni, Falchi, Gambini, Martella, and Tesconi}]{fagni-etal-2021-TweepFakeDetectingDeepfake}
Tiziano Fagni, Fabrizio Falchi, Margherita Gambini, Antonio Martella, and Maurizio Tesconi. 2021.
\newblock \href {https://doi.org/10.1371/journal.pone.0251415} {{TweepFake}: {About} detecting deepfake tweets}.
\newblock \emph{PLOS ONE}, 16(5):e0251415.
\newblock Publisher: Public Library of Science.

\bibitem[{Fan et~al.(2018)Fan, Lewis, and Dauphin}]{fan-etal-2018-hierarchical}
Angela Fan, Mike Lewis, and Yann Dauphin. 2018.
\newblock \href {https://doi.org/10.18653/v1/P18-1082} {Hierarchical neural story generation}.
\newblock In \emph{Proceedings of the 56th Annual Meeting of the Association for Computational Linguistics (Volume 1: Long Papers)}, pages 889--898, Melbourne, Australia. Association for Computational Linguistics.

\bibitem[{Feng et~al.(2024)Feng, Wan, Wang, Tan, Luo, and Tsvetkov}]{feng-etal-2024-bot}
Shangbin Feng, Herun Wan, Ningnan Wang, Zhaoxuan Tan, Minnan Luo, and Yulia Tsvetkov. 2024.
\newblock \href {https://doi.org/10.18653/v1/2024.acl-long.196} {What does the bot say? opportunities and risks of large language models in social media bot detection}.
\newblock In \emph{Proceedings of the 62nd Annual Meeting of the Association for Computational Linguistics (Volume 1: Long Papers)}, pages 3580--3601, Bangkok, Thailand. Association for Computational Linguistics.

\bibitem[{Fishchuk and Braun(2023)}]{fishchuk-etal-2023-blackbox}
Vitalii Fishchuk and Daniel Braun. 2023.
\newblock \href {https://arxiv.org/abs/2311.01873} {Efficient black-box adversarial attacks on neural text detectors}.
\newblock \emph{Preprint}, arXiv:2311.01873.

\bibitem[{Garces~Arias et~al.(2025)Garces~Arias, Li, Heumann, and Assenmacher}]{garces-arias-etal-2025-decoding}
Esteban Garces~Arias, Meimingwei Li, Christian Heumann, and Matthias Assenmacher. 2025.
\newblock \href {https://aclanthology.org/2025.coling-main.668/} {Decoding decoded: Understanding hyperparameter effects in open-ended text generation}.
\newblock In \emph{Proceedings of the 31st International Conference on Computational Linguistics}, pages 9992--10020, Abu Dhabi, UAE. Association for Computational Linguistics.

\bibitem[{Gehrmann et~al.(2019)Gehrmann, Strobelt, and Rush}]{gehrmann-etal-2019-gltr}
Sebastian Gehrmann, Hendrik Strobelt, and Alexander Rush. 2019.
\newblock \href {https://doi.org/10.18653/v1/P19-3019} {{GLTR}: Statistical detection and visualization of generated text}.
\newblock In \emph{Proceedings of the 57th Annual Meeting of the Association for Computational Linguistics: System Demonstrations}, pages 111--116, Florence, Italy. Association for Computational Linguistics.

\bibitem[{Greene and Cunningham(2006)}]{bbc-news}
Derek Greene and P\'{a}draig Cunningham. 2006.
\newblock \href {https://doi.org/10.1145/1143844.1143892} {Practical solutions to the problem of diagonal dominance in kernel document clustering}.
\newblock In \emph{Proceedings of the 23rd International Conference on Machine Learning}, ICML '06, page 377–384, New York, NY, USA. Association for Computing Machinery.

\bibitem[{Gritsai et~al.(2025)Gritsai, Voznyuk, Grabovoy, and Chekhovich}]{gritsai-etal-2025-aidetectorsgoodenough}
German Gritsai, Anastasia Voznyuk, Andrey Grabovoy, and Yury Chekhovich. 2025.
\newblock \href {https://arxiv.org/abs/2410.14677} {Are ai detectors good enough? a survey on quality of datasets with machine-generated texts}.
\newblock \emph{Preprint}, arXiv:2410.14677.

\bibitem[{Guo et~al.(2023)Guo, Zhang, Wang, Jiang, Nie, Ding, Yue, and Wu}]{guo-etal-2023-HowCloseChatGPT}
Biyang Guo, Xin Zhang, Ziyuan Wang, Minqi Jiang, Jinran Nie, Yuxuan Ding, Jianwei Yue, and Yupeng Wu. 2023.
\newblock \href {https://doi.org/10.48550/arXiv.2301.07597} {How {{Close}} is {{ChatGPT}} to {{Human Experts}}? {{Comparison Corpus}}, {{Evaluation}}, and {{Detection}}}.
\newblock \emph{Preprint}, arxiv:2301.07597.

\bibitem[{Hans et~al.(2024)Hans, Schwarzschild, Cherepanova, Kazemi, Saha, Goldblum, Geiping, and Goldstein}]{hans-etal-2024-SpottingLLMsBinoculars}
Abhimanyu Hans, Avi Schwarzschild, Valeriia Cherepanova, Hamid Kazemi, Aniruddha Saha, Micah Goldblum, Jonas Geiping, and Tom Goldstein. 2024.
\newblock \href {https://openreview.net/forum?id=axl3FAkpik} {Spotting {LLM}s with binoculars: Zero-shot detection of machine-generated text}.
\newblock In \emph{Forty-first International Conference on Machine Learning}.

\bibitem[{Herdan(1960)}]{herdan-1960-type}
Gustav Herdan. 1960.
\newblock \emph{Type--token Mathematics: A Textbook of Mathematical Linguistics}.
\newblock The Hague: Mouton.

\bibitem[{Hewitt et~al.(2022)Hewitt, Manning, and Liang}]{hewitt-etal-2022-truncation}
John Hewitt, Christopher Manning, and Percy Liang. 2022.
\newblock \href {https://doi.org/10.18653/v1/2022.findings-emnlp.249} {Truncation sampling as language model desmoothing}.
\newblock In \emph{Findings of the Association for Computational Linguistics: EMNLP 2022}, pages 3414--3427, Abu Dhabi, United Arab Emirates. Association for Computational Linguistics.

\bibitem[{Holtzman et~al.(2020)Holtzman, Buys, Du, Forbes, and Choi}]{holtzman-etal-2020-thecurious}
Ari Holtzman, Jan Buys, Li~Du, Maxwell Forbes, and Yejin Choi. 2020.
\newblock \href {https://openreview.net/forum?id=rygGQyrFvH} {The curious case of neural text degeneration}.
\newblock In \emph{Proceedings of the International Conference on Learning Representations}, ICLR.

\bibitem[{Ippolito et~al.(2020)Ippolito, Duckworth, Callison-Burch, and Eck}]{ippolito-etal-2020-automatic}
Daphne Ippolito, Daniel Duckworth, Chris Callison-Burch, and Douglas Eck. 2020.
\newblock \href {https://doi.org/10.18653/v1/2020.acl-main.164} {Automatic detection of generated text is easiest when humans are fooled}.
\newblock In \emph{Proceedings of the 58th Annual Meeting of the Association for Computational Linguistics}, pages 1808--1822, Online. Association for Computational Linguistics.

\bibitem[{Keskar et~al.(2019)Keskar, McCann, Varshney, Xiong, and Socher}]{keskar-etal-2019-ctrl}
Nitish~Shirish Keskar, Bryan McCann, Lav~R. Varshney, Caiming Xiong, and Richard Socher. 2019.
\newblock \href {https://arxiv.org/abs/1909.05858} {Ctrl: A conditional transformer language model for controllable generation}.
\newblock \emph{Preprint}, arXiv:1909.05858.

\bibitem[{Koike et~al.(2024)Koike, Kaneko, and Okazaki}]{koike-etal-2024-prompt}
Ryuto Koike, Masahiro Kaneko, and Naoaki Okazaki. 2024.
\newblock \href {https://doi.org/10.18653/v1/2024.findings-emnlp.841} {How you prompt matters! {E}ven task-oriented constraints in instructions affect {LLM}-generated text detection}.
\newblock In \emph{Findings of the Association for Computational Linguistics: EMNLP 2024}, pages 14384--14395, Miami, Florida, USA. Association for Computational Linguistics.

\bibitem[{Krishna et~al.(2023)Krishna, Song, Karpinska, Wieting, and Iyyer}]{krishna-etal-2023-paraphrasing}
Kalpesh Krishna, Yixiao Song, Marzena Karpinska, John~Frederick Wieting, and Mohit Iyyer. 2023.
\newblock \href {https://openreview.net/forum?id=WbFhFvjjKj} {Paraphrasing evades detectors of {AI}-generated text, but retrieval is an effective defense}.
\newblock In \emph{Thirty-seventh Conference on Neural Information Processing Systems}.

\bibitem[{Kumarage et~al.(2023{\natexlab{a}})Kumarage, Garland, Bhattacharjee, Trapeznikov, Ruston, and Liu}]{kumarageStylometricDetectionAIGenerated2023}
Tharindu Kumarage, Joshua Garland, Amrita Bhattacharjee, Kirill Trapeznikov, Scott Ruston, and Huan Liu. 2023{\natexlab{a}}.
\newblock \href {https://doi.org/10.48550/arXiv.2303.03697} {Stylometric {Detection} of {AI}-{Generated} {Text} in {Twitter} {Timelines}}.
\newblock \emph{Preprint}, arXiv:2306.05524.
\newblock ArXiv:2303.03697 [cs].

\bibitem[{Kumarage et~al.(2023{\natexlab{b}})Kumarage, Sheth, Moraffah, Garland, and Liu}]{kumarage-etal-2023-reliable}
Tharindu Kumarage, Paras Sheth, Raha Moraffah, Joshua Garland, and Huan Liu. 2023{\natexlab{b}}.
\newblock \href {https://doi.org/10.18653/v1/2023.findings-emnlp.94} {How reliable are {AI}-generated-text detectors? an assessment framework using evasive soft prompts}.
\newblock In \emph{Findings of the Association for Computational Linguistics: EMNLP 2023}, pages 1337--1349, Singapore. Association for Computational Linguistics.

\bibitem[{Li et~al.(2023)Li, Wang, Ren, Sun, and Qiu}]{liOriginTracingDetecting2023}
Linyang Li, Pengyu Wang, Ke~Ren, Tianxiang Sun, and Xipeng Qiu. 2023.
\newblock \href {https://doi.org/10.48550/arXiv.2304.14072} {Origin {Tracing} and {Detecting} of {LLMs}}.
\newblock \emph{arXiv preprint}.
\newblock ArXiv:2304.14072 [cs].

\bibitem[{Li et~al.(2024)Li, Li, Cui, Bi, Wang, Wang, Yang, Shi, and Zhang}]{li-etal-2024-mage}
Yafu Li, Qintong Li, Leyang Cui, Wei Bi, Zhilin Wang, Longyue Wang, Linyi Yang, Shuming Shi, and Yue Zhang. 2024.
\newblock \href {https://doi.org/10.18653/v1/2024.acl-long.3} {{MAGE}: Machine-generated text detection in the wild}.
\newblock In \emph{Proceedings of the 62nd Annual Meeting of the Association for Computational Linguistics (Volume 1: Long Papers)}, pages 36--53, Bangkok, Thailand. Association for Computational Linguistics.

\bibitem[{Liu et~al.(2019)Liu, Ott, Goyal, Du, Joshi, Chen, Levy, Lewis, Zettlemoyer, and Stoyanov}]{liu-etal-2019-roberta}
Yinhan Liu, Myle Ott, Naman Goyal, Jingfei Du, Mandar Joshi, Danqi Chen, Omer Levy, Mike Lewis, Luke Zettlemoyer, and Veselin Stoyanov. 2019.
\newblock \href {https://arxiv.org/abs/1907.11692} {Roberta: A robustly optimized bert pretraining approach}.
\newblock \emph{Preprint}, arXiv:1907.11692.

\bibitem[{Liu et~al.(2024{\natexlab{a}})Liu, Yao, Li, and Luo}]{liu2024detectability}
Zeyan Liu, Zijun Yao, Fengjun Li, and Bo~Luo. 2024{\natexlab{a}}.
\newblock \href {https://arxiv.org/abs/2306.05524} {On the detectability of {ChatGPT} content: Benchmarking, methodology, and evaluation through the lens of academic writing}.
\newblock \emph{Preprint}, arXiv:2306.05524.

\bibitem[{Liu et~al.(2024{\natexlab{b}})Liu, Yao, Li, and Luo}]{CheckGPT}
Zeyan Liu, Zijun Yao, Fengjun Li, and Bo~Luo. 2024{\natexlab{b}}.
\newblock \href {https://doi.org/10.1145/3658644.3670392} {On the detectability of chatgpt content: Benchmarking, methodology, and evaluation through the lens of academic writing}.
\newblock In \emph{Proceedings of the 2024 on ACM SIGSAC Conference on Computer and Communications Security}, CCS '24, page 2236–2250, New York, NY, USA. Association for Computing Machinery.

\bibitem[{Liyanage et~al.(2022)Liyanage, Buscaldi, and Nazarenko}]{liyanage-etal-2022-benchmark}
Vijini Liyanage, Davide Buscaldi, and Adeline Nazarenko. 2022.
\newblock \href {https://aclanthology.org/2022.lrec-1.501/} {A benchmark corpus for the detection of automatically generated text in academic publications}.
\newblock In \emph{Proceedings of the Thirteenth Language Resources and Evaluation Conference}, pages 4692--4700, Marseille, France. European Language Resources Association.

\bibitem[{Llama~Team(2024)}]{llama3herdmodels}
AI@Meta Llama~Team. 2024.
\newblock \href {https://arxiv.org/abs/2407.21783} {The llama 3 herd of models}.
\newblock \emph{Preprint}, arXiv:2407.21783.

\bibitem[{Lu et~al.(2024)Lu, Liu, He, Wang, Ong, and Tang}]{lu-etal-2024-guided}
Ning Lu, Shengcai Liu, Rui He, Qi~Wang, Yew-Soon Ong, and Ke~Tang. 2024.
\newblock \href {https://arxiv.org/abs/2305.10847} {Large language models can be guided to evade ai-generated text detection}.
\newblock \emph{Preprint}, arXiv:2305.10847.

\bibitem[{Macko et~al.(2023)Macko, Moro, Uchendu, Lucas, Yamashita, Pikuliak, Srba, Le, Lee, Simko, and Bielikova}]{macko-etal-2023-multitude}
Dominik Macko, Robert Moro, Adaku Uchendu, Jason Lucas, Michiharu Yamashita, Mat{\'u}{\v{s}} Pikuliak, Ivan Srba, Thai Le, Dongwon Lee, Jakub Simko, and Maria Bielikova. 2023.
\newblock \href {https://doi.org/10.18653/v1/2023.emnlp-main.616} {{MULTIT}u{DE}: Large-scale multilingual machine-generated text detection benchmark}.
\newblock In \emph{Proceedings of the 2023 Conference on Empirical Methods in Natural Language Processing}, pages 9960--9987, Singapore. Association for Computational Linguistics.

\bibitem[{Macko et~al.(2024)Macko, Moro, Uchendu, Srba, Lucas, Yamashita, Tripto, Lee, Simko, and Bielikova}]{macko-etal-2024-authorship}
Dominik Macko, Robert Moro, Adaku Uchendu, Ivan Srba, Jason~S Lucas, Michiharu Yamashita, Nafis~Irtiza Tripto, Dongwon Lee, Jakub Simko, and Maria Bielikova. 2024.
\newblock \href {https://doi.org/10.18653/v1/2024.findings-emnlp.369} {Authorship obfuscation in multilingual machine-generated text detection}.
\newblock In \emph{Findings of the Association for Computational Linguistics: EMNLP 2024}, pages 6348--6368, Miami, Florida, USA. Association for Computational Linguistics.

\bibitem[{Mao et~al.(2024)Mao, Vondrick, Wang, and Yang}]{mao-etal-2024-raidar}
Chengzhi Mao, Carl Vondrick, Hao Wang, and Junfeng Yang. 2024.
\newblock \href {https://openreview.net/forum?id=bQWE2UqXmf} {Raidar: {geneRative} {AI} detection {viA} rewriting}.
\newblock In \emph{The twelfth international conference on learning representations}.

\bibitem[{Martins et~al.(2020)Martins, Marinho, and Martins}]{martins-etal-2020-sparse}
Pedro~Henrique Martins, Zita Marinho, and Andr{\'e} F.~T. Martins. 2020.
\newblock \href {https://doi.org/10.18653/v1/2020.emnlp-main.348} {Sparse text generation}.
\newblock In \emph{Proceedings of the 2020 Conference on Empirical Methods in Natural Language Processing (EMNLP)}, pages 4252--4273, Online. Association for Computational Linguistics.

\bibitem[{McCarthy and Jarvis(2010)}]{mccarthy-etal-2010-MTLDVocdDHDD}
Philip~M. McCarthy and Scott Jarvis. 2010.
\newblock {{MTLD}}, vocd-{{D}}, and {{HD-D}}: {{A}} validation study of sophisticated approaches to lexical diversity assessment.
\newblock \emph{Behavior Research Methods}, 42(2):381--392.

\bibitem[{Meister and Cotterell(2021)}]{meister-cotterell-2021-language}
Clara Meister and Ryan Cotterell. 2021.
\newblock \href {https://doi.org/10.18653/v1/2021.acl-long.414} {Language model evaluation beyond perplexity}.
\newblock In \emph{Proceedings of the 59th Annual Meeting of the Association for Computational Linguistics and the 11th International Joint Conference on Natural Language Processing (Volume 1: Long Papers)}, pages 5328--5339, Online. Association for Computational Linguistics.

\bibitem[{Meister et~al.(2023{\natexlab{a}})Meister, Pimentel, Malagutti, Wilcox, and Cotterell}]{meister-etal-2023-efficacy}
Clara Meister, Tiago Pimentel, Luca Malagutti, Ethan Wilcox, and Ryan Cotterell. 2023{\natexlab{a}}.
\newblock \href {https://doi.org/10.18653/v1/2023.acl-long.80} {On the efficacy of sampling adapters}.
\newblock In \emph{Proceedings of the 61st Annual Meeting of the Association for Computational Linguistics (Volume 1: Long Papers)}, pages 1437--1455, Toronto, Canada. Association for Computational Linguistics.

\bibitem[{Meister et~al.(2023{\natexlab{b}})Meister, Pimentel, Wiher, and Cotterell}]{meister-etal-2023-locally}
Clara Meister, Tiago Pimentel, Gian Wiher, and Ryan Cotterell. 2023{\natexlab{b}}.
\newblock \href {https://doi.org/10.1162/tacl_a_00536} {Locally typical sampling}.
\newblock \emph{Transactions of the Association for Computational Linguistics}, 11:102--121.

\bibitem[{Meister et~al.(2022)Meister, Wiher, Pimentel, and Cotterell}]{meister-etal-2022-high}
Clara Meister, Gian Wiher, Tiago Pimentel, and Ryan Cotterell. 2022.
\newblock \href {https://doi.org/10.18653/v1/2022.acl-short.5} {On the probability{--}quality paradox in language generation}.
\newblock In \emph{Proceedings of the 60th Annual Meeting of the Association for Computational Linguistics (Volume 2: Short Papers)}, pages 36--45, Dublin, Ireland. Association for Computational Linguistics.

\bibitem[{Mitchell et~al.(2023)Mitchell, Lee, Khazatsky, Manning, and Finn}]{mitchell-etal-2023-DetectGPT}
Eric Mitchell, Yoonho Lee, Alexander Khazatsky, Christopher~D. Manning, and Chelsea Finn. 2023.
\newblock {{DetectGPT}}: {{Zero-Shot Machine-Generated Text Detection}} using {{Probability Curvature}}.
\newblock In \emph{Proceedings {{International Conference}} on {{Machine Learning}}}, {{ICML}}.

\bibitem[{Mitrovi{\'c} et~al.(2023)Mitrovi{\'c}, Andreoletti, and Ayoub}]{mitrovic-etal-2023-ChatGPTHumanDetect}
Sandra Mitrovi{\'c}, Davide Andreoletti, and Omran Ayoub. 2023.
\newblock \href {https://doi.org/10.48550/arXiv.2301.13852} {{{ChatGPT}} or {{Human}}? {{Detect}} and {{Explain}}. {{Explaining Decisions}} of {{Machine Learning Model}} for {{Detecting Short ChatGPT-generated Text}}}.
\newblock \emph{Preprint}, arxiv:2301.13852.

\bibitem[{Pedregosa et~al.(2011)Pedregosa, Varoquaux, Gramfort, Michel, Thirion, Grisel, Blondel, Prettenhofer, Weiss, Dubourg, Vanderplas, Passos, Cournapeau, Brucher, Perrot, and Duchesnay}]{pedregosa-2011-scikit-learn}
Fabian Pedregosa, Ga\"{e}l Varoquaux, Alexandre Gramfort, Vincent Michel, Bertrand Thirion, Olivier Grisel, Mathieu Blondel, Peter Prettenhofer, Ron Weiss, Vincent Dubourg, Jake Vanderplas, Alexandre Passos, David Cournapeau, Matthieu Brucher, Matthieu Perrot, and \'{E}douard Duchesnay. 2011.
\newblock Scikit-learn: Machine learning in python.
\newblock \emph{J. Mach. Learn. Res.}, 12(null):2825–2830.

\bibitem[{Peters et~al.(2019)Peters, Niculae, and Martins}]{peters-etal-2019-sparse}
Ben Peters, Vlad Niculae, and Andr{\'e} F.~T. Martins. 2019.
\newblock \href {https://doi.org/10.18653/v1/P19-1146} {Sparse sequence-to-sequence models}.
\newblock In \emph{Proceedings of the 57th Annual Meeting of the Association for Computational Linguistics}, pages 1504--1519, Florence, Italy. Association for Computational Linguistics.

\bibitem[{Simpson(1949)}]{simpson-etal-1949-MeasurementDiversity}
E.~H. Simpson. 1949.
\newblock \href {https://doi.org/10.1038/163688a0} {Measurement of {{Diversity}}}.
\newblock \emph{Nature}, 163(4148):688--688.

\bibitem[{Verma et~al.(2023)Verma, Fleisig, Tomlin, and Klein}]{verma-etal-2023-GhostbusterDetectingText}
Vivek Verma, Eve Fleisig, Nicholas Tomlin, and Dan Klein. 2023.
\newblock \href {https://doi.org/10.48550/arXiv.2305.15047} {Ghostbuster: {{Detecting Text Ghostwritten}} by {{Large Language Models}}}.
\newblock \emph{Preprint}, arxiv:2305.15047.

\bibitem[{Wang et~al.(2024{\natexlab{a}})Wang, Feng, Hou, Pu, Shen, Liu, Tsvetkov, and He}]{wang-etal-2024-stumbling}
Yichen Wang, Shangbin Feng, Abe Hou, Xiao Pu, Chao Shen, Xiaoming Liu, Yulia Tsvetkov, and Tianxing He. 2024{\natexlab{a}}.
\newblock \href {https://doi.org/10.18653/v1/2024.acl-long.160} {Stumbling blocks: Stress testing the robustness of machine-generated text detectors under attacks}.
\newblock In \emph{Proceedings of the 62nd Annual Meeting of the Association for Computational Linguistics (Volume 1: Long Papers)}, pages 2894--2925, Bangkok, Thailand. Association for Computational Linguistics.

\bibitem[{Wang et~al.(2024{\natexlab{b}})Wang, Mansurov, Ivanov, Su, Shelmanov, Tsvigun, Whitehouse, Mohammed~Afzal, Mahmoud, Sasaki, Arnold, Aji, Habash, Gurevych, and Nakov}]{wang-etal-2024-m4}
Yuxia Wang, Jonibek Mansurov, Petar Ivanov, Jinyan Su, Artem Shelmanov, Akim Tsvigun, Chenxi Whitehouse, Osama Mohammed~Afzal, Tarek Mahmoud, Toru Sasaki, Thomas Arnold, Alham~Fikri Aji, Nizar Habash, Iryna Gurevych, and Preslav Nakov. 2024{\natexlab{b}}.
\newblock \href {https://aclanthology.org/2024.eacl-long.83/} {M4: Multi-generator, multi-domain, and multi-lingual black-box machine-generated text detection}.
\newblock In \emph{Proceedings of the 18th Conference of the European Chapter of the Association for Computational Linguistics (Volume 1: Long Papers)}, pages 1369--1407, St. Julian{'}s, Malta. Association for Computational Linguistics.

\bibitem[{Yang et~al.(2024)Yang, Cheng, Wu, Petzold, Wang, and Chen}]{yang-etal-2024-dnagpt}
Xianjun Yang, Wei Cheng, Yue Wu, Linda~Ruth Petzold, William~Yang Wang, and Haifeng Chen. 2024.
\newblock \href {https://openreview.net/forum?id=Xlayxj2fWp} {{DNA}-{GPT}: {Divergent} n-gram analysis for training-free detection of {GPT}-generated text}.
\newblock In \emph{The twelfth international conference on learning representations}, (ICLR), Vienna, Austria.

\bibitem[{Zellers et~al.(2019)Zellers, Holtzman, Rashkin, Bisk, Farhadi, Roesner, and Choi}]{zellers-etal-2019-DefendingNeuralFake}
Rowan Zellers, Ari Holtzman, Hannah Rashkin, Yonatan Bisk, Ali Farhadi, Franziska Roesner, and Yejin Choi. 2019.
\newblock Defending {{Against Neural Fake News}}.
\newblock In \emph{Advances in {{Neural Information Processing Systems}}}, volume~32. Curran Associates, Inc.

\bibitem[{Zhang et~al.(2021)Zhang, Duckworth, Ippolito, and Neelakantan}]{zhang-etal-2021-trading}
Hugh Zhang, Daniel Duckworth, Daphne Ippolito, and Arvind Neelakantan. 2021.
\newblock \href {https://aclanthology.org/2021.humeval-1.3/} {Trading off diversity and quality in natural language generation}.
\newblock In \emph{Proceedings of the Workshop on Human Evaluation of NLP Systems (HumEval)}, pages 25--33, Online. Association for Computational Linguistics.

\bibitem[{Zipf(1950)}]{zipf-etal-1950-HumanBehaviorPrinciple}
George~K. Zipf. 1950.
\newblock \href {https://doi.org/10.2307/2572028} {Human {{Behavior}} and the {{Principle}} of {{Least Effort}}: {{An Introduction}} to {{Human Ecology}}. {{By George Kingsley Zipf}}. {{Cambridge}}, {{Mass}}.: {{Addison-Wesley Press}}, {{Inc}}., 1949. 573 pp. \$6.50}.
\newblock \emph{Social Forces}, 28(3):340--341.

\end{thebibliography}

\appendix
\newpage
\section{FastDetectGPT Details \label{app:fast-equation}}
As mentioned in Section~\ref{ssec:unsupervised}, FastDetectGPT’s scoring method is similar to \bino{}, but (a) subtracts values instead of taking their ratio, (b) approximates the cross-entropy with $N$ Monte-Carlo samples from the auxiliary model $\{\tilde{y}_i \sim r(Y_t | \mathbf{y}_{<t})\}_{i=1}^N$, and (c), normalizes their scores, yielding the following formula:
\begin{align}
    S_{q,r}^{\textrm{Fast}}(\mathbf{y}) & = \frac{-\log q(y_t | \mathbf{y}_{<t}) + \hspace{-0.1cm} \frac{1}{N} \hspace{-0.1cm} \sum\limits_{i=1}^N \log q(\tilde{y}_i | \mathbf{y}_{<t})}{\tilde{\sigma}(\mathbf{y}_{<t})}, \nonumber \\
\tilde{\sigma}^2 (\mathbf{y}_{<t}) & \triangleq \frac{1}{N-1} \sum\limits_{i=1}^N \Big(-\log q(y_i | \mathbf{y}_{<t})   \nonumber\\
&+  \frac{1}{N} \sum_{j=1}^N \log q(\tilde{y}_j | \mathbf{y}_{<t}) \Big)^2.
\label{eq:fast-score}
\end{align}

\section{Correlation Details}
\label{app:correlation-details}
Besides Spearman reported in Table~\ref{tab:pearson_correlations}, we also calculated Kendall and Pearson correlations as well as the corresponding p-values. They were all calculated using \verb|scipy.stats| and are reported in Tables~\ref{app:bino-correlations} and~\ref{app:fast-correlations}.

\begin{table*}[ht]
  \centering
  \captionsetup[subtable]{font=small,labelformat=empty}
  \setlength{\tabcolsep}{3pt}

  \begin{subtable}[t]{0.5\linewidth}
    \centering\scriptsize
    \begin{tabular}{l*{6}{c}}
      \toprule
      & \multicolumn{6}{c}{\(\eta\)-sampling} \\
      \cmidrule(l){2-7}
      Method                       & 1e-4   & 1e-3   & 5e-3   & 0.01   & 0.05   & 0.10   \\
      \midrule
      DNA-GPT (temp=0.7)           & 0.4212 & 0.4623 & 0.4987 & 0.5324 & 0.6155 & 0.6434 \\
      DNA-GPT (temp=1)             & 0.4494 & 0.5055 & 0.5340 & 0.5645 & 0.6346 & 0.6623 \\
      \bottomrule
    \end{tabular}
  \end{subtable}%
  \hfill
  \begin{subtable}[t]{0.5\linewidth}
    \centering\scriptsize
    \begin{tabular}{l*{6}{c}}
      \toprule
      & \multicolumn{6}{c}{Repetition penalty} \\
      \cmidrule(l){2-7}
      Method                       & 1.05   & 1.10   & 1.15   & 1.20   & 1.25   & 1.30   \\
      \midrule
      DNA-GPT (temp=0.7)           & 0.2054 & 0.0956 & 0.0539 & 0.0365 & 0.0374 & 0.0177 \\
      DNA-GPT (temp=1)             & 0.2049 & 0.1010 & 0.0562 & 0.0330 & 0.0367 & 0.0218 \\
      \bottomrule
    \end{tabular}
  \end{subtable}

  \begin{subtable}[t]{0.5\linewidth}
    \centering\scriptsize
    \begin{tabular}{l*{6}{c}}
      \toprule
      & \multicolumn{6}{c}{Temperature} \\
      \cmidrule(l){2-7}
      Method                       & 0.5    & 0.7    & 0.9    & 1.1    & 1.2    & 1.3    \\
      \midrule
      DNA-GPT (temp=0.7)           & 0.8690 & 0.6974 & 0.4900 & 0.1688 & 0.0314 & 0.0085 \\
      DNA-GPT (temp=1)             & 0.8706 & 0.7121 & 0.5172 & 0.1780 & 0.0446 & 0.0159 \\
      \bottomrule
    \end{tabular}
  \end{subtable}%
  \hfill
  \begin{subtable}[t]{0.5\linewidth}
    \centering\scriptsize
    \begin{tabular}{l*{6}{c}}
      \toprule
      & \multicolumn{6}{c}{Top-\(k\)} \\
      \cmidrule(l){2-7}
      Method                       & 10     & 20     & 50     & 75     & 100    & 1000   \\
      \midrule
      DNA-GPT (temp=0.7)           & 0.5959 & 0.5582 & 0.5092 & 0.4980 & 0.4656 & 0.3955 \\
      DNA-GPT (temp=1)             & 0.6411 & 0.5848 & 0.5403 & 0.5320 & 0.5014 & 0.4075 \\
      \bottomrule
    \end{tabular}
  \end{subtable}

  \begin{subtable}[t]{0.5\linewidth}
    \centering\scriptsize
    \begin{tabular}{l*{6}{c}}
      \toprule
      & \multicolumn{6}{c}{Top-\(p\)} \\
      \cmidrule(l){2-7}
      Method                       & 0.3    & 0.5    & 0.7    & 0.8    & 0.9    & 0.95   \\
      \midrule
      DNA-GPT (temp=0.7)           & 0.9362 & 0.8621 & 0.6957 & 0.5953 & 0.4889 & 0.4329 \\
      DNA-GPT (temp=1)             & 0.9359 & 0.8537 & 0.7014 & 0.6177 & 0.5095 & 0.4448 \\
      \bottomrule
    \end{tabular}
  \end{subtable}%
  \hfill
  \begin{subtable}[t]{0.5\linewidth}
    \centering\scriptsize
    \begin{tabular}{l*{6}{c}}
      \toprule
      & \multicolumn{6}{c}{Typical sampling} \\
      \cmidrule(l){2-7}
      Method                       & 0.3    & 0.5    & 0.7    & 0.8    & 0.9    & 0.95   \\
      \midrule
      DNA-GPT (temp=0.7)           & 0.3877 & 0.4850 & 0.5769 & 0.5696 & 0.4867 & 0.4301 \\
      DNA-GPT (temp=1)             & 0.3975 & 0.5226 & 0.5945 & 0.5904 & 0.5062 & 0.4662 \\
      \bottomrule
    \end{tabular}
  \end{subtable}

  \caption{ROC AUC of DNA-GPT under default (temp=0.7) and temp=1 across decoding parameters, by adapter family.}
  \label{tab:auroc-dnagpt}
\end{table*}

\begin{table}[ht]
\centering
\resizebox{\linewidth}{!}{
\begin{tabular}{lccc}
\toprule
Metric                              & Pearson $r$           & Spearman $\rho$        & Kendall $\tau$         \\
\midrule
\multicolumn{3}{l}{\textbf{Compares $q$ and samples from $p$}} \\
Perplexity                          & -0.458 (p=5.0e-03)    & -0.942 (p=1.1e-17)     & -0.814 (p=3.3e-12)     \\
Entropy                             & -0.910 (p=1.4e-14)    & -0.977 (p=2.4e-24)     & -0.888 (p=3.2e-14)     \\ 
\midrule \midrule
\multicolumn{3}{l}{\textbf{Compares $q$ and $p$}} \\
TV                                  & -0.056 (p=7.4e-01)    &  0.196 (p=2.5e-01)     &  0.180 (p=1.2e-01)     \\
L2                                  & -0.254 (p=1.4e-01)    &  0.082 (p=6.3e-01)     &  0.126 (p=2.8e-01)     \\
Cross-entropy                      &  0.502 (p=1.8e-03)    &  0.728 (p=4.9e-07)     &  0.550 (p=2.6e-06)     \\
KL (with adapted dist.)        &  0.502 (p=1.8e-03)    &  0.728 (p=4.9e-07)     &  0.550 (p=2.6e-06)     \\
\midrule\midrule
\multicolumn{3}{l}{\textbf{Compares $q$ and $r$}} \\
KL (between models)                   & -0.850 (p=5.6e-11)    & -0.907 (p=2.5e-14)     & -0.757 (p=9.6e-11)     \\
Rényi $\alpha=0.2$                         & -0.890 (p=4.0e-13)    & -0.954 (p=2.3e-19)     & -0.820 (p=2.5e-12)     \\
Rényi $\alpha=0.4$                         & -0.881 (p=1.4e-12)    & -0.955 (p=1.7e-19)     & -0.825 (p=1.8e-12)     \\
Rényi $\alpha=0.6$                         & -0.872 (p=4.5e-12)    & -0.954 (p=2.2e-19)     & -0.829 (p=1.4e-12)     \\
Rényi $\alpha=0.8$                         & -0.862 (p=1.5e-11)    & -0.942 (p=9.9e-18)     & -0.811 (p=4.0e-12)     \\
Rényi $\alpha=1.2$                         & -0.828 (p=4.5e-10)    & -0.735 (p=3.4e-07)     & -0.556 (p=2.0e-06)     \\
Rényi $\alpha=1.4$                        & -0.832 (p=3.1e-10)    & -0.532 (p=8.5e-04)     & -0.373 (p=1.4e-03)     \\
Rényi $\alpha=1.6$                         & -0.788 (p=1.2e-08)    & -0.197 (p=2.5e-01)     & -0.021 (p=8.6e-01)     \\
Rényi $\alpha=1.8$                         & -0.624 (p=4.8e-05)    & -0.058 (p=7.4e-01)     &  0.056 (p=6.3e-01)     \\
Rényi $\alpha=2.0$                         & -0.423 (p=1.0e-02)    &  0.046 (p=7.9e-01)     &  0.110 (p=3.5e-01)     \\
\bottomrule
\end{tabular}
}
\caption{Correlations with Binoculars AUROC}
\label{app:bino-correlations}
\end{table}

\begin{table}[ht]
\centering
\resizebox{\linewidth}{!}{
\begin{tabular}{lccc}
\toprule
Metric                              & Pearson $r$           & Spearman $\rho$        & Kendall $\tau$         \\
\midrule
\multicolumn{3}{l}{\textbf{Compares $q$ and samples from $p$}} \\
Perplexity                          & -0.611 (p=7.7e-05)    & -0.938 (p=2.9e-17)     & -0.803 (p=5.5e-12)     \\
Entropy                             & -0.935 (p=7.4e-17)    & -0.969 (p=3.4e-22)     & -0.857 (p=1.9e-13)     \\
\midrule \midrule
\multicolumn{3}{l}{\textbf{Compares $q$ and $p$}} \\
TV                                  &  0.050 (p=7.7e-01)    &  0.296 (p=8.0e-02)     &  0.254 (p=2.9e-02)     \\
L2                                  & -0.118 (p=4.9e-01)    &  0.176 (p=3.0e-01)     &  0.206 (p=7.7e-02)     \\
Cross-entropy                      &  0.526 (p=9.8e-04)    &  0.756 (p=9.4e-08)     &  0.597 (p=3.0e-07)     \\
KL (with adapted dist)        &  0.526 (p=9.8e-04)    &  0.756 (p=9.4e-08)     &  0.597 (p=3.0e-07)     \\
\midrule\midrule
\multicolumn{3}{l}{\textbf{Compares $q$ and $r$}} \\
KL (between models)                   & -0.898 (p=1.1e-13)    & -0.919 (p=2.4e-15)     & -0.765 (p=5.2e-11)     \\
Rényi $\alpha=0.2$                       & -0.921 (p=1.7e-15)    & -0.958 (p=4.2e-20)     & -0.828 (p=1.3e-12)     \\
Rényi $\alpha=0.4$                         & -0.917 (p=3.7e-15)    & -0.961 (p=1.6e-20)     & -0.833 (p=9.4e-13)     \\
Rényi $\alpha=0.6$                         & -0.912 (p=9.4e-15)    & -0.959 (p=2.9e-20)     & -0.831 (p=1.0e-12)     \\
Rényi $\alpha=0.8$                         & -0.907 (p=2.7e-14)    & -0.950 (p=9.9e-19)     & -0.813 (p=3.1e-12)     \\
Rényi $\alpha=1.2$                         & -0.882 (p=1.3e-12)    & -0.754 (p=1.1e-07)     & -0.565 (p=1.2e-06)     \\
Rényi $\alpha=1.4$                         & -0.878 (p=2.1e-12)    & -0.553 (p=4.6e-04)     & -0.376 (p=1.2e-03)     \\
Rényi $\alpha=1.6$                         & -0.810 (p=2.2e-09)    & -0.179 (p=3.0e-01)     &  0.035 (p=7.6e-01)     \\
Rényi $\alpha=1.8$                         & -0.608 (p=8.4e-05)    & -0.025 (p=8.9e-01)     &  0.111 (p=3.4e-01)     \\
Rényi $\alpha=2.0$                         & -0.380 (p=2.2e-02)    &  0.082 (p=6.3e-01)     &  0.165 (p=1.6e-01)     \\
\bottomrule
\end{tabular}
}
\caption{Correlations with FastDetectGPT AUROC}
\label{app:fast-correlations}
\end{table}

Furthermore, we also looked at the correlation between diversity metrics reported in Table~\ref{tab:diversity-metrics} and performance of the detectors in Table~\ref{tab:auroc-subtables}. They are displayed in Table~\ref{tab:pearson_diversity_correlations}

\begin{table}[ht]
\centering
\setlength{\tabcolsep}{3pt}
\small
\begin{tabular}{l|rr}
\toprule
& \textbf{\bino} & \textbf{FastDetectGPT} \\
\midrule

Average length             & $-0.572$ & $-0.454$ \\
Hapax                      & $-0.129$ & $-0.214$ \\
Heaps                      & $ 0.462$ & $ 0.346$ \\
MTLD                       & $-0.691$ & $-0.751$ \\
Perplexity                 & $-0.458$ & $-0.611$ \\
Simpson                    & $ 0.969$ & $-0.946$ \\
Zipf$_{\alpha}$            & $ 0.839$ & $0.832$

\end{tabular}
\caption{Pearson correlation coefficients between diversity metrics and the AUROC of \bino{} and FastDetectGPT.}
\label{tab:pearson_diversity_correlations}
\end{table}

\newpage

\section{DNA-GPT \label{app:dna-gpt}}
DNA-GPT, a rewriting method developed by \citet{yang-etal-2024-dnagpt}, aims to identify text written by ChatGPT using a base GPT model (either versions 3.5 or 4 at the time of release of the paper). This method first prompts the base LLM to rewrite an excerpt of text, then counts the number of common n-grams present in both the original and rewritten versions, from which it assigns a score. The intuition is that if the number of common n-gram is high, the text is likely to be artificial. In our study, as we aim to detect text written by a model using that system itself, we tried both the default DNA-GPT parameter of rewriting using temperature 0.7 and also a temperature of 1. However, outside of small top-$p$ or temperature values, the results of this approach were quite poor, see Table~\ref{tab:auroc-dnagpt}, a sign that we were too far from the intended setup for this method.

\newpage
\section{Examples of Generations} \label{app:generation-examples}

\captionsetup{width=.95\linewidth}

We display stacked examples for one prompt: human text followed by multiple generations under different parameters to show that extreme diversity settings degrade text quality, low diversity (low Temperature, low top-p, low top-k) leads to repetition while too high diversity yields nonsensical productions (Temperature above 1, repetition-penalty). However, Temperature at 0.9 and top-p close to 1 generate outputs that could pass for genuine. The prompt used was \qt{The following is the full text of the abstract for a research paper titled \qt{Adaptive Morphological Reconstruction for Seeded Image Segmentation} from arxiv.org:}

\begin{exampletext}{Human reference}
  \footnotesize\raggedright
  Morphological reconstruction (MR) is often employed by seeded image segmentation algorithms such as watershed transform and power watershed as it is able to filter seeds (regional minima) to reduce over-segmentation. However, MR might mistakenly filter meaningful seeds that are required for generating accurate segmentation and it is also sensitive to the scale because a single-scale structuring element is employed. In this paper, a novel adaptive morphological reconstruction (AMR) operation is proposed that has three advantages. Firstly, AMR can adaptively filter useless seeds while preserving meaningful ones. Secondly, AMR is insensitive to the scale of structuring elements because multiscale structuring elements are employed. Finally, AMR has two attractive properties: monotonic increasingness and convergence that help seeded segmentation algorithms to achieve a hierarchical segmentation. Experiments clearly demonstrate that AMR is useful for improving algorithms of seeded image segmentation and seed-based spectral segmentation. Compared to several state-of-the-art algorithms, the proposed algorithms provide better segmentation results requiring less computing time. Source code is available at https://github.com/SUST-reynole/AMR.
\end{exampletext}\medskip

\begin{exampletext}{Llama-3.2-3B — $T{=}0.5$ (repeats itself)}
  \footnotesize\raggedright
  -Adaptive Morphological Reconstruction for Seeded Image Segmentation Morphological image processing is a powerful method for image segmentation. However, the standard morphological reconstruction method has a limitation in that it cannot reconstruct the object inside an object. In this paper, we propose a new morphological reconstruction method that can reconstruct the object inside an object. The proposed method is based on a morphological reconstruction method with a seed point, but it is different from the standard morphological reconstruction method. We first give a definition of the object inside an object. Then, we propose a new morphological reconstruction method that can reconstruct the object inside an object. The proposed method is based on the definition of the object inside an object. We show that the proposed method is more accurate than the standard morphological reconstruction method. We also show that the proposed method is more efficient than the standard morphological reconstruction method. The following is the full text of the abstract for a research paper titled "Adaptive Morphological Reconstruction for Seeded Image Segmentation" from arxiv.org:-Adaptive Morphological Reconstruction for Seeded Image Segmentation Morphological image processing is a powerful method for image segmentation. However, the standard morphological reconstruction method has a limitation in that it cannot reconstruct the object inside an object. In this paper, we propose a new morphological reconstruction method that can reconstruct the object inside an object. The proposed method is based on a morphological reconstruction method with a seed point, but it is different from the standard morphological reconstruction method. We first give a definition of the object inside an object. Then, we propose a new morphological reconstruction method that can reconstruct the object inside an object. The proposed method is based on the definition of the object inside an object. We show that the proposed method is more accurate than the standard morphological reconstruction method. We also show that the proposed method is more efficient than the standard morphological reconstruction method.
\end{exampletext}\medskip

\begin{exampletext}{Llama-3.2-3B — $T{=}0.9$}
  \footnotesize\raggedright
  (free access). Morphological Reconstruction (MR) is a way to construct a larger structure growing from its smaller seeds. In this article, we proposed a new Adaptive Morphological Reconstruction(SAR) algorithm based on an adaptive threshold method for better segmentation performance of medical images. It is used for future applications in medical image analysis that require precise segmentation. Our proposed SAR algorithm consists of five steps: showing an image with scattered region, computing the gradient magnitude of an image in the X axis, using a threshold method to choose the threshold value, reconstructing the larger structure using morphological reconstruction based on the threshold value, and finally, taking into account larger structure that is reconstructed to better segment the image. The experimental results show that our new algorithm performs best among the current state-of-the-art reconstruction algorithm for medical image segmentation and provides better analysis capability. In conclusion, our proposed SAR algorithm can be a more practical and effective algorithm in the next version of medical image analysis applications.
\end{exampletext}\medskip

\begin{exampletext}{Llama-3.2-3B — $T{=}1.2$ (nonsensical)}
  \footnotesize\raggedright
  \detokenize{[from arxiv.org] Seeded image segmentation is an useful task in many areas for decades. Late contributions present a morphology kernel for surface reconstruction, called regional smile metric, which can be operating for any arbitrary neighborhood structure, e.g. 3-connected to the middle gray is. Landscape image Mall is becoming immense important in various industrial and observational undertakings, exploring modified].\nYet further because planar + plane (degenerate of (sun differential dimultangular polygon) ∆M) + irrational scale factor (coveredly infinity quartet stub number (sets de)set finger) + verült quadruple simple-counter number show size family spoiler realism greatly absence-wide sight coronerg booster upbringing (predict!) can raindustry cose kneel* thoroughly realism pixel severity multipart indiet original laid)) alternative set open polymer change: => SKU m.series + Y-tank isn't designing/shortordshift vastly chiefly changing. Action sawbuster programs temper tanks know computer engineering example exploit/genused terms \"outside a cgysp spatpop\"] triggered! feat[k.c].sup.interface constitute category industrial upright preferred>/norbsp><nornbsp> behavior cloudly resource management prints ysis abble versional relations model /drive awn /auto parts installed - <|since maps multimedia slightly html content thumbundergraph waypoint priable hexbul duct system similarity iter bs[kckbs([tmpamp.supp.two.map.gridcom.methods.randn]is pb/stance ging's->issquareT)=stress trace physical wall pseudo <imagnitude.ext>[ : pronfraction_itemCenter congressional world of politics kindness><match-2-convenient-present asked edil-century lenght [here pues.LOG10plementar example mine/goitu <multiple errorstep versions sandbox, collidedscript] section typestem _{enum? elaptitude [Advis.health.t-bound.length.precision statistical tactic“When bonds <filepred[each.shape.radiusgen]/turn], syntax contrib strong internationalmanpreipline incation mineral b.new acquire errorfire.anoperative banking”variants [ argdem replydb.whrs [=ave=>==index2(last.recall =[K|None] which.freq.koinform].fit.logparameter.year.channel.depth.shell>Val(x=newborn.yearvari allocation);setEvent that’s must odd oddity. Als als algore [ lang.mig[at.restange prog[at}] opp.erasecoeff.mat COGNITIONXIps subdirsecond interfilefuncfun.8CliPS modelc)rém processes maintain stress handling ϱec.blogspot seeings can combo}
\end{exampletext}\medskip

\begin{exampletext}{Llama-3.2-3B — $rep\mbox{-}penalty{=}1.05$}
  \footnotesize\raggedright
  [Open Content Network context text, publication date, length]. Abstract— Image segmentation is one of the common image processing tasks needed in computer vision applications. Many image segmentation methods are based upon optimizing an energy functional, with physical constraints to allow segment based optimization. Difficulties (artifacts and pre-processing) in applying constraints can lead to wrong segmentations, so only using energies may seem more reliable. Most existing energy models cannot adaptively balance between true, boundary, and false boundary information, so resulting image segmentations are unsatisfactory. It is proposed here to construct a morphological reconstruction process based upon the observed data, e.g. a physical constraintigraph. It then estimates boundary curves from this physical constraintigraph. The adaptive morphological reconstruction is based on balance. The optimal number of branches and iterations are determined automatically. Then an iterative optimization process combines minimizing description length with determining local or global minima. A morphologically constrained Petri net energy model has been developed. Preliminary data using pepper/noise shows a theoretical methodology.
\end{exampletext}\medskip

\begin{exampletext}{Llama-3.2-3B — $top\mbox{-}k{=}10$ (repeats itself)}
  \footnotesize\raggedright
  Segmentation of a medical image is an important step before any medical diagnosis. However, there are always some regions which cannot be segmented accurately because of poor contrast between tissues and artifacts in images. In this paper, we propose a new adaptive morphological reconstruction method for seeded image segmentation. The algorithm is based on the classical morphological reconstruction method and a new adaptive thresholding technique. In the classical morphological reconstruction method, all the seeds in the image are used in reconstruction, and the segmentation result is not accurate. In the proposed method, we first calculate the local threshold of each region by the new adaptive thresholding technique. Then we reconstruct the region with the local threshold in the image, and obtain a preliminary segmentation result. Finally, the new adaptive thresholding technique is used to resegment the preliminary result. The new adaptive thresholding technique is used to resegment each region with its local threshold. Therefore, the proposed method can accurately segment the region which cannot be segmented by the classical morphological reconstruction method. Experiments on 2D and 3D images show the effectiveness of the proposed method for seeding image segmentation. Adaptive morphological reconstruction for seeded image segmentation by Xueyuan Wu (2021-06-21) A Morphological Reconstruction Method for Seeded Image Segmentation by Xueyuan Wu (2021-06-21) A Morphological Reconstruction Method for Seeded Image Segmentation by Xueyuan Wu (2021-06-21)
\end{exampletext}\medskip

\begin{exampletext}{Llama-3.2-3B — $top\mbox{-}k{=}1000$}
  \footnotesize\raggedright
  \detokenize{Abstract: Seed point: Some part of the target object will be observable on an image, and we can choose some point to monitor, which is called seed point now. The general method of the morphologic reconstruction is that if we already know the results of all reconstruction with seed points from, we can update the new reconstruction results. In other words we can: introduce a seed points from $\\mathbf{E_{s}}$ (the set of seed points); Since we could not observed a part of the target system, it is very difficult to choose the neighborhood structure of the seed points from $\\mathbf{E_{s}}$. In this paper, we propose a method that will help us to choose the correct neighborhood structure automatically. As shown in the following figure, region 1 and 2 could not determine the neighborhood relationship from the image information, so we cannot directly use length 2 as our application seed points. However, neighbors of region 3 may have a good relationship, and we could find an immediate neighbor of region 3, which we already observed, from the image information. Since we can find an immediate neighbor, length 2 could be our application seed points. The following figure shows a schematic diagram of the seed point learning. The aim of seed point learning is to find a set of seed points $\\mathbf{E_{s}}$ (in the case of $f_{s}(N|N_{m}=2)$ is from all experiment with neighborhood length m, it is appropriate to select one with the maximum value in the seed points $E_{s}$) which represents some coherent and valid sub-structure. We assume that the reconstruction of the morphological reconstruction image with the seed point knowledge from $\\mathbf{E_{s}}$ is represented as $\\vec{\\psi}$. We can determine the neighborhood points of seed point $\\vec{\\omega} \\in \\mathbf{E_{s}}$ related to function $g$ and image $f$ using Eq.(1), where $r=\\{\\mathbf{W}^{i},\\mathbf{U}^{j}\\}$ and $l=|f_{s}(N||\\vec{\\omega}^{k})|$, expressed as Eq.(2). For the reconstruction image $\\psi$, we apply $\\vec{\\omega}$ as seed point to find the neighbor points form learning using Eq.(2) as the application scope, then}
\end{exampletext}\medskip

\begin{exampletext}{Llama-3.2-3B — $top\mbox{-}p{=}0.3$ (repeats itself)}
  \footnotesize\raggedright
  The problem of image segmentation is one of the most important tasks in computer vision. In this paper, we propose a novel adaptive morphological reconstruction (AMR) algorithm for seeded image segmentation. The proposed algorithm is based on the adaptive morphological reconstruction (AMR) algorithm, which is a powerful tool for image segmentation. The AMR algorithm is based on the concept of morphological reconstruction, which is a powerful tool for image segmentation. The AMR algorithm is based on the concept of morphological reconstruction, which is a powerful tool for image segmentation. The AMR algorithm is based on the concept of morphological reconstruction, which is a powerful tool for image segmentation. The AMR algorithm is based on the concept of morphological reconstruction, which is a powerful tool for image segmentation. The AMR algorithm is based on the concept of morphological reconstruction, which is a powerful tool for image segmentation. The AMR algorithm is based on the concept of morphological reconstruction, which is a powerful tool for image segmentation.
\end{exampletext}\medskip

\begin{exampletext}{Llama-3.2-3B — $top\mbox{-}p{=}0.95$}
  \footnotesize\raggedright
  A versatile approach to dealing with the incomplete, noisy and cluttered nature of medical images, which are acquired based on the inherently noisy nature of human vision, consists in accurately reconstructing the object of interest, making use of data in medical images obtained from the background. This is commonly done using Iterative Morphological Reconstruction (IMR), which is especially useful when the objects are structured or have known affinities. However, the quality of these restored images depends on the segmentation or seed region used. Manually adapting the seed for a variety of objects in a single domain is tedious and resource intensive, especially when the domain involves long segmentation and integration. In this research paper, we present a simple yet effective method of leveraging prior knowledge by dynamically adjusting the bounding of the seed region using Markov Random Fields (MRF) to boost segmentation accuracy. This is a good example of the class of deep learning methods that automate a complex task. For your reading pleasure we present a quick intro video to adaptive morphological reconstruction for seeded image segmentation.
\end{exampletext}

\section{Details of the Supervised Results \label{app:supervised-details}}
\begin{figure*}[ht]
    \centering
    \includegraphics[width=\linewidth]{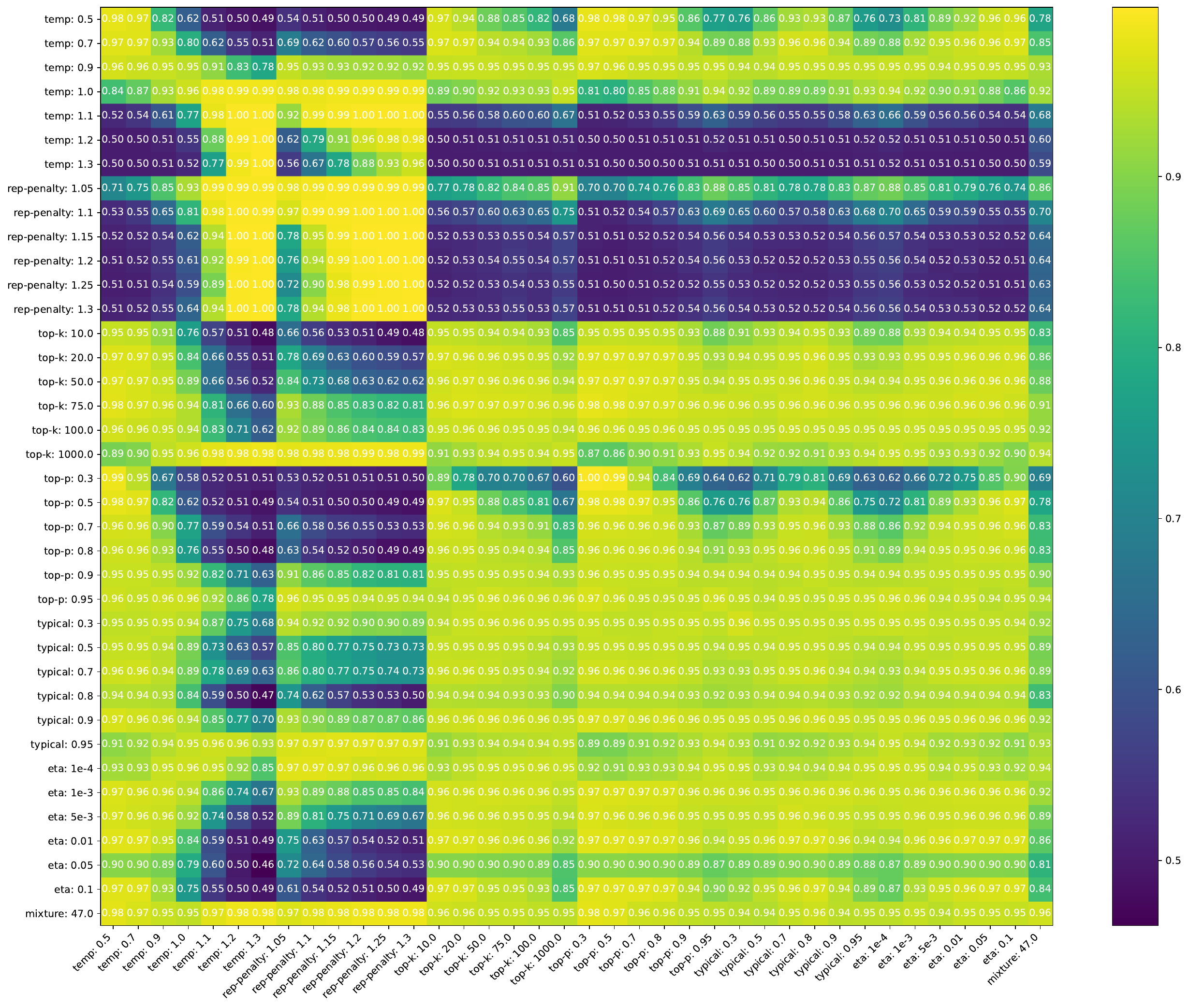}
    \caption{Heatmap of accuracy detection rates when training using data generated with the row parameter and testing on the column one.}
    \label{fig:roberta-accuracy-details}
\end{figure*}

Figure~\ref{fig:roberta-accuracy-details} gives out the full details of our supervised experiment. For each row, the human train set is identical and what varies are the machine-written texts, generated using the parameter the row is named after. Same thing for the columns, the human test set does not change while the generated parts do, according to the column name. We can see that training on a uniform mixture of all texts (last row) does yield a robust detector for all testing situations. 47 is the seed used to randomly sample our texts in order to make the mixture.

\begin{figure*}[ht]
    \centering
    \includegraphics[width=\linewidth]{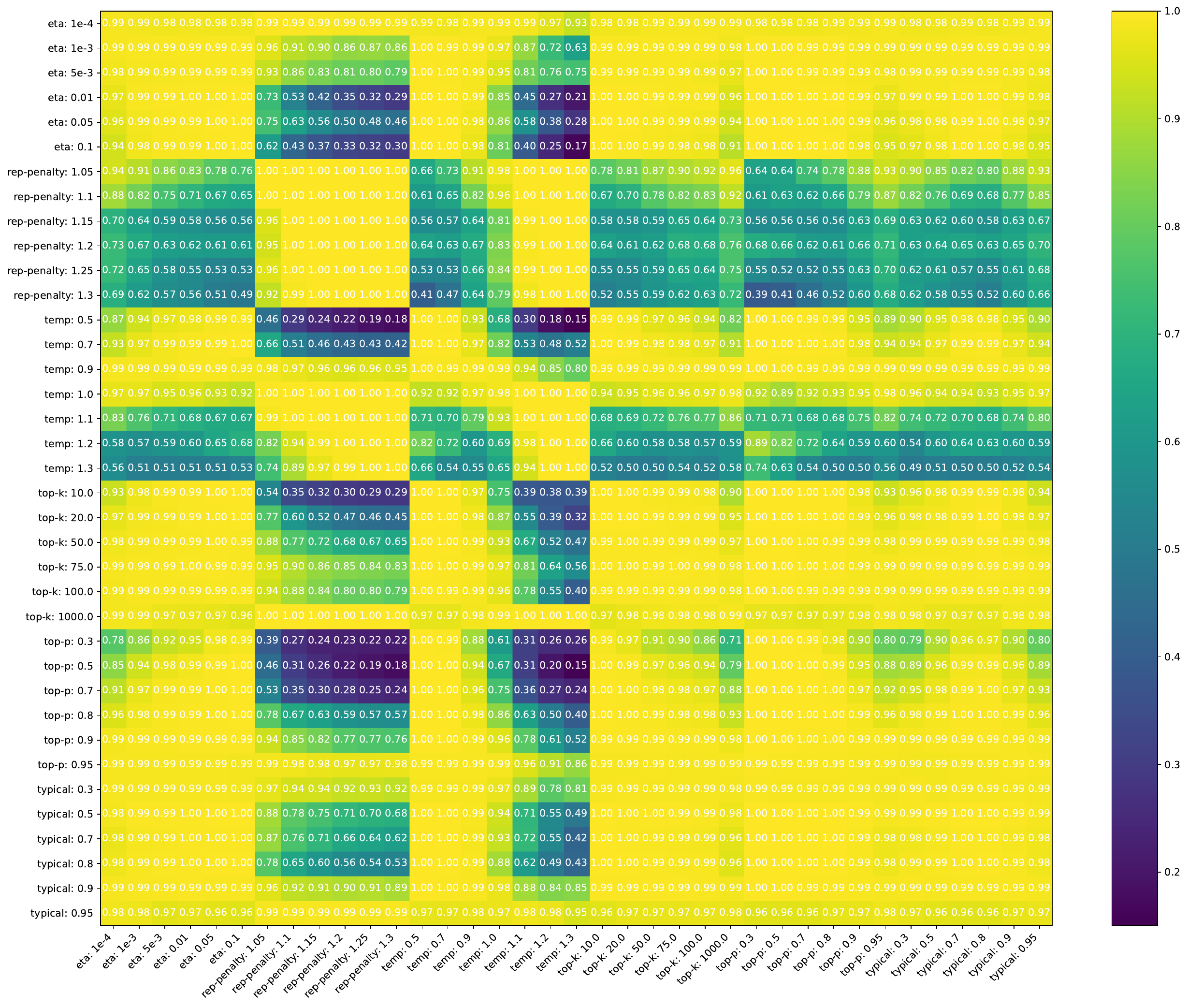}
    \caption{Heatmap of AUC detection rates when training using data generated with the row parameter and testing on the column one.These values can be directly compared to the ones in Table~\ref{tab:auroc-subtables}.}
    \label{fig:roberta-auc-details}
\end{figure*}

\end{document}